\documentclass[journal]{IEEEtran}
\usepackage[T1]{fontenc}
\usepackage{amsmath,amssymb,amsfonts}
\usepackage{bm}
\usepackage{graphicx}
\usepackage{adjustbox}
\usepackage{booktabs}
\usepackage{multirow}
\usepackage[table]{xcolor}
\usepackage{tikz}
\usepackage{pgfplots}
\pgfplotsset{compat=1.17}
\usetikzlibrary{positioning,arrows.meta,fit,backgrounds,calc}
\usepackage{algorithm}
\usepackage{algpseudocode}
\usepackage{cite}
\usepackage[hidelinks]{hyperref}

\definecolor{cphys}{HTML}{0072B2}
\definecolor{cnn}{HTML}{E69F00}
\definecolor{cnnd}{HTML}{A06A00}
\definecolor{cgreen}{HTML}{009E73}
\definecolor{cred}{HTML}{D55E00}
\definecolor{cpoly}{HTML}{CC79A7}
\definecolor{csky}{HTML}{56B4E9}
\definecolor{good}{HTML}{D9F0E3}

\newcommand{\phys}{\textsc{Phys}}
\newcommand{\physfc}{\textsc{Phys-fc}}
\newcommand{\poly}[1]{\textsc{Poly#1}}
\newcommand{\nn}[1]{\textsc{NN-#1}}
\newcommand{\hct}{\textsc{HCTide}}
\newcommand{\Mtwo}{M_2}

\pgfplotsset{
  hct/.style={
    tick label style={font=\scriptsize},
    label style={font=\footnotesize},
    legend style={font=\tiny, draw=none, fill=white, fill opacity=0.85, text opacity=1},
    grid style={gray!18},
    axis line style={gray!60!black},
  }
}

\graphicspath{{figs/}}

\newcommand{\figorbox}[2]{%
  \IfFileExists{figs/#2.pdf}{\includegraphics[width=#1]{#2.pdf}}{%
    \IfFileExists{figs/#2.png}{\includegraphics[width=#1]{#2.png}}{%
      \IfFileExists{figs/#2.jpg}{\includegraphics[width=#1]{#2.jpg}}{%
        \fbox{\parbox[c][3.5cm][c]{\dimexpr#1-2\fboxsep-2\fboxrule\relax}{\centering\footnotesize
          Missing figure: \texttt{figs/\detokenize{#2}}}}}}}}

\begin{document}

\title{HCTide: A Hard-Constrained Differentiable Framework for Learning and Auditing Tidal Dissipation Closures}

\author{Boris~Kriuk and Tanisha~Kiratsata%
\thanks{B. Kriuk is with the Department of Information Systems, Business Statistics and Operations Management, The Hong Kong University of Science and Technology, Clear Water Bay, Kowloon, Hong Kong SAR, China (e-mail: bkriuk@connect.ust.hk).}%
\thanks{T. Kiratsata is with the Department of Mathematics and Statistics and the Department of Physics and Astronomy, Trent University, 1600 W Bank Dr, Peterborough, ON, Canada (e-mail: tanishakiratsata@trentu.ca).}}

\markboth{}%
{Kriuk and Kiratsata: HCTide, Hard-Constrained Learning of Tidal Dissipation Closures}

\maketitle

\begin{abstract}
Dissipation closures are the main source of error in forward global tide models; learning them from data requires exact physics and evaluation of transferable skill. We present \hct{}, to our knowledge the first framework that trains tidal dissipation closures, from physical laws to neural networks, through a round-off-exact global tidal solve and audits them with by-products of that same solve. The closure predicts only a drag field and a self-attraction-and-loading factor; the frequency-domain tidal equations are solved on an energy-consistent C-grid. Every parameter vector yields a dissipative, non-singular operator, discrete work equals discrete dissipation to $6\times10^{-16}$, and an implicit adjoint that reuses the forward factorization gives exact discrete gradients whose cost rises only 15\% from 4 to 17\,154 parameters. A five-axis admissibility audit (spatial extrapolation, frequency transfer, energy, interpretability, projection stability), an altimetric-atlas benchmark and a grid-refinement test run in about 100\,min on a laptop CPU. On 226 UHSLC tide gauges, every constrained closure beats Gaussian-process interpolation by 17--29\%. A 114-parameter neural closure leads a 4-parameter physical law by 3.3\,cm under random cross-validation but trails it by 4.2\,cm under spatial blocking; the audit links this penalty to a hemispherically asymmetric drag correction at northern shelf-sea latitudes. Replacing $\sin\varphi$ by $|\sin\varphi|$ yields the best random-CV closure, statistically level with the physical law under blocking. Refinement to $1^\circ$ is consistent with the audit: the signed-latitude network loses its lead, while the symmetric one stays level. A global energy constraint brings the neural closure within 1\% of the 2.4\,TW altimetric target without losing skill relative to the physical law.
\end{abstract}


\section{Introduction}
\label{sec:intro}

\IEEEPARstart{T}{idal} predictions underpin coastal flood forecasting, navigation, tidal-energy siting and the planning of coastal infrastructure under sea-level rise~\cite{pugh2014,haigh2020}. Data-assimilative atlases are highly accurate in the open ocean~\cite{egbert2002,stammer2014,lyard2021}. However, \emph{forward} tide models are still needed wherever assimilation is impossible: for past or future bathymetry and sea level~\cite{pickering2017}, inside coupled ocean--climate simulations, and in regions without altimetric coverage. In forward models the dominant source of error is not the dynamics but the \emph{dissipation closure}, i.e., how bottom friction and the conversion of barotropic energy into internal waves are parameterized~\cite{arbic2004,jayne2001,green2013,buijsman2015}. These closures carry a handful of hand-tuned coefficients and are calibrated with ad hoc procedures.

Machine learning offers a way to learn such closures from observations~\cite{rasp2018,bolton2019,zanna2020,frezat2022}. Two requirements must be met before learned closures can enter operational geophysical inverse problems. First, physics must hold exactly. Physics-informed neural networks (PINNs) add the PDE residual to the loss as a \emph{soft} penalty~\cite{raissi2019,karniadakis2021}, so conservation laws hold only approximately, the optimization is ill-conditioned~\cite{wang2021,krishnapriyan2021}, and energy-producing closures are not excluded. Second, evaluation must measure \emph{transferable} skill. Tide gauges are strongly clustered along coastlines, so random splits reward interpolation~\cite{roberts2017,ploton2020,valavi2019}, and a closure's physical credibility shows only when it is used to \emph{project}, for instance under sea-level rise.

\begin{figure*}[t]
\centering
\resizebox{\textwidth}{!}{%
\begin{tikzpicture}[
  font=\sffamily\small,
  >={Stealth[length=2.2mm]},
  neur/.style={circle, draw=#1!75!black, fill=#1!25, minimum size=4.6mm, inner sep=0pt, line width=0.5pt},
  conn/.style={draw=gray!50, line width=0.3pt},
  box/.style={draw=#1!70!black, fill=#1!10, rounded corners=3pt, align=center,
              text width=2.7cm, minimum height=2.4cm, line width=0.6pt},
  aux/.style={draw=gray!60, fill=white, rounded corners=2pt, align=center, dashed,
              text width=2.7cm, minimum height=0.7cm, font=\sffamily\footnotesize},
  chip/.style={draw=cgreen!65!black, fill=white, rounded corners=2pt, align=center,
               text width=3.1cm, minimum height=0.95cm, font=\sffamily\footnotesize},
  lyr/.style={font=\sffamily\scriptsize, text=gray!35!black},
  fwd/.style={->, line width=0.9pt, draw=black!70},
  adj/.style={->, line width=0.9pt, draw=cred, dashed},
]
\foreach \i/\lab in {1/{$H$},2/{$|\nabla H|$},3/{$\varphi$}}{
  \node[neur=gray] (I\i) at (0,{(2-\i)*0.95}) {};
  \node[anchor=east, font=\sffamily\footnotesize] (L\i) at ([xshift=-1.2mm]I\i.west) {\lab};
}
\foreach \j in {1,...,5}{
  \node[neur=cnn] (A\j) at (1.7,{(3-\j)*0.62}) {};
  \node[neur=cnn] (B\j) at (3.4,{(3-\j)*0.62}) {};
}
\node[neur=cred, font=\sffamily\scriptsize] (O) at (5.1,0) {$h$};
\foreach \i in {1,2,3}{\foreach \j in {1,...,5}{\draw[conn] (I\i) -- (A\j);}}
\foreach \i in {1,...,5}{\foreach \j in {1,...,5}{\draw[conn] (A\i) -- (B\j);}}
\foreach \j in {1,...,5}{\draw[conn] (B\j) -- (O);}
\node[lyr] (l0) at (0,-1.75)   {input $\bm\phi(\mathbf{x})$};
\node[lyr] (l1) at (1.7,-1.75) {hidden ($\tanh$)};
\node[lyr] (l2) at (3.4,-1.75) {hidden ($\tanh$)};
\node[lyr] (l3) at (5.1,-1.75) {raw drag};
\begin{scope}[on background layer]
\node[fit=(L1)(L2)(L3)(I1)(A1)(A5)(O)(l0)(l3), draw=cnn!70!black, fill=cnn!6,
      rounded corners=4pt, inner sep=6pt,
      label={[font=\sffamily\small\bfseries, text=cnnd, anchor=south]north:Learnable closure $g_\theta$ (MLP shown)}] (nnbox) {};
\end{scope}

\node[box=cnn, right=0.9cm of nnbox] (cons) {\textbf{Positivity layer}\\[3pt]
  $r=r_{\mathrm{ref}}\,e^{s\tanh(h/s)}$\\[2pt] $\kappa=r/H+\kappa_{\min}>0$};
\node[box=cphys, right=0.9cm of cons] (solv) {\textbf{Differentiable tidal solver}\\[3pt]
  $\mathbf{A}(\theta;\omega_k)\,\mathbf{x}_k=\mathbf{b}_k$\\[2pt] exact sparse LU\\ energy-consistent C-grid};
\node[box=cgreen, right=0.9cm of solv] (loss) {\textbf{Gauge misfit}\\[3pt]
  $\mathcal{L}(\theta)=$\\ $\sum_{g,k}w_g\,|\eta_{gk}-\hat Z_{gk}|^2$};

\draw[fwd] (O) -- (O -| cons.west);
\draw[fwd] (cons) -- (solv);
\draw[fwd] (solv) -- (loss);

\node[aux, above=0.6cm of solv] (forc) {astronomical forcing $\eta^{\mathrm{eq}}_k$};
\node[aux, above=0.6cm of loss] (obs)  {tide-gauge harmonics $\hat Z_{gk}$};
\draw[fwd] (forc) -- (solv);
\draw[fwd] (obs) -- (loss);

\coordinate (ybot) at ($(nnbox.south)+(0,-0.6)$);
\draw[adj] (loss.south) -- (loss.south |- ybot)
  -- node[fill=white, inner sep=2pt, font=\sffamily\footnotesize, text=cred!80!black]
     {implicit adjoint $\mathbf{A}_k^{H}\bm\lambda_k=\partial_{\mathbf{x}_k}\mathcal{L}$ (reuses LU factors) $\;\rightarrow\;\nabla_\theta\mathcal{L}$}
  (nnbox.south |- ybot) -- (nnbox.south);

\coordinate (audtl) at ($(nnbox.west |- ybot)+(0,-0.55)$);
\node[anchor=north west, font=\sffamily\small\bfseries, text=cgreen!50!black, inner sep=0pt]
  (audtitle) at ($(audtl)+(0.15,0)$) {Admissibility audit of every fitted $\theta^\star$ (by-products of the same solve)};
\node[chip, anchor=north west] (a1) at ($(audtitle.south west)+(0,-0.2)$) {\textbf{A1}\\ spatial-block CV};
\node[chip, right=0.42cm of a1] (a2) {\textbf{A2}\\ frequency transfer};
\node[chip, right=0.42cm of a2] (a3) {\textbf{A3}\\ energy budget};
\node[chip, right=0.42cm of a3] (a4) {\textbf{A4}\\ distillation};
\node[chip, right=0.42cm of a4] (a5) {\textbf{A5}\\ sea-level projection};
\begin{scope}[on background layer]
\node[fit=(audtitle)(a1)(a5), draw=cgreen!60!black, fill=cgreen!6, rounded corners=4pt, inner sep=5pt] {};
\end{scope}
\end{tikzpicture}}
\caption{The \hct{} framework. A learnable closure $g_\theta$ maps fixed bathymetric and geographic features to a raw drag $h$. An MLP is shown; the physical law and the polynomial closures plug into the same slot. A positivity layer turns $h$ into a bounded, strictly positive damping rate. For each constituent the discretized tidal equations are then solved to round-off by sparse LU, and only the gauge misfit touches the data. Solid arrows: forward pass. Dashed red arrow: implicit adjoint, which reuses the LU factors, so the gradient cost does not depend on $\dim\theta$; L-BFGS then updates $\theta$. Every fitted $\theta^\star$ is audited along five axes.}
\label{fig:arch}
\end{figure*}
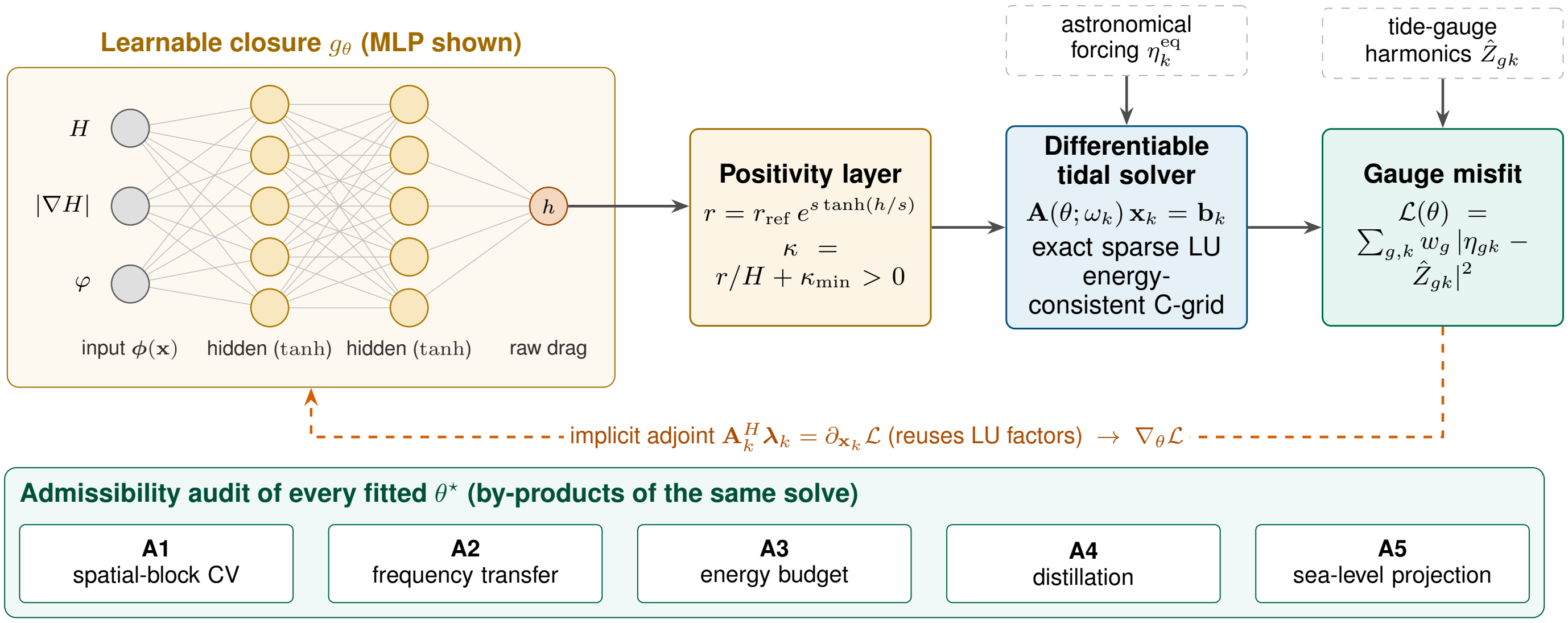

\hct{} (Fig.~\ref{fig:arch}) meets both requirements in one framework. It is a \emph{solver-in-the-loop}~\cite{um2020} model in which the learnable component is confined to the uncertain physics: a positive drag field $r_{\mathrm{eff}}(\mathbf{x};\theta)$ and a scalar self-attraction-and-loading (SAL) factor $\beta$. Everything else (gravity waves, Coriolis, continuity and astronomical forcing) is discretized and solved \emph{to round-off} as a sparse linear system per tidal constituent. Physical admissibility is therefore a property of the architecture rather than of the loss. Gradients come from an implicit-function adjoint~\cite{amos2017,bai2019,blondel2022} that reuses the forward LU factorization, so a 17\,154-parameter network trains at almost the cost of a 4-parameter physical law. The same solve yields energy budgets, dissipation maps and perturbed-geometry projections, so \hct{} doubles as an \emph{audit engine}: any closure, physical or learned, can be checked on five independent axes in minutes, placed on an absolute scale against an altimetric atlas, and re-tested on a refined grid. Its distinguishing strength is verifiability: every claim about a learned closure, from energy closure to transferable skill, is checked by the same direct solve, and the verdict the audit reaches at $2^\circ$ is consistent with what refinement to $1^\circ$ shows.

\textbf{Contributions.}
\begin{enumerate}
\item \emph{Framework.} A hard-constrained, differentiable tide model that combines a bounded positivity layer, an energy-consistent discretization and an implicit adjoint layer behind a single plug-in closure interface. For \emph{every} parameter vector the operator is non-singular and dissipative, and discrete work equals discrete dissipation to round-off. The gradient costs one adjoint back-substitution per constituent, independent of $\dim\theta$ (Sec.~\ref{sec:method}).
\item \emph{Verification.} Energy closure is $\le6.1\times10^{-16}$ and adjoint/finite-difference agreement $\le7.8\times10^{-7}$ for five closures. Cost per evaluation grows by only 15\% over a $4\,300\times$ increase in parameters. Refined to $1^\circ$ (114\,524 unknowns), every fitted model still closes the energy budget to $3.3\times10^{-16}$ (Secs.~\ref{sec:res:verif} and~\ref{sec:res:res}).
\item \emph{Admissibility audit and external benchmark.} A reproducible five-axis protocol built on the framework's by-products: spatial extrapolation (random vs.\ spherical-block CV), frequency transfer, energy admissibility, interpretability by distillation, and projection stability under sea-level rise. It is complemented by an atlas benchmark that separates the grid-representativeness error common to all models from closure error, both at the gauges and over the global field, and by a grid-refinement test that checks whether a learned gain survives a change of resolution. The full suite of 170 fits takes about 100\,min of optimizer time on a laptop CPU, and the refinement test adds 3\,min (Sec.~\ref{sec:setup}).
\item \emph{Findings that guide closure design.} (i)~The physics solver adds value: every closure beats Gaussian-process (GP) interpolation, the physical closure's margin \emph{grows} under extrapolation, and the closures remove 33--55\% of the GP's RMS excess over the representativeness floor. (ii)~Spatial-block validation separates transferable from region-specific skill, a distinction random CV cannot make, and gauge stratification locates the region-specific part at coastal gauges. (iii)~The learned drag acts as a diagnostic: it points to missing dissipation in the latitude bands of the northern shelf seas. (iv)~Frequency transfer, energy budgets and fold-ensemble projections quantify properties that held-out skill alone does not reveal. (v)~The framework converts these diagnostics into more robust closures: a hemispherically symmetric latitude feature yields the best random-CV closure, which is also level with the physical closure under spatial blocking, and a global energy constraint brings the neural closure onto the altimetric budget target without loss of held-out skill relative to the physical law. (vi)~Grid refinement is consistent with the audit: at $1^\circ$ the representativeness floor falls by 4.6\,cm and the physical closure's error by 4.3\,cm, the signed-latitude network's random-CV lead is no longer present, and the symmetric network remains level with the physical closure, as spatial-block validation at $2^\circ$ had indicated (Sec.~\ref{sec:results}).
\end{enumerate}

The practical outcome is a decision procedure. With the few hundred stations typical of global gauge networks, a compact physical closure is the recommended baseline, a neural closure is a diagnostic that points to deficient physics, and a learned closure is promoted once its gains hold on all five audit axes. The framework makes each axis cheap to evaluate. The hemispherically symmetric network studied here is the only learned closure that is best under random CV at $2^\circ$ and statistically level with the physical closure both under spatial blocking and after refinement to $1^\circ$.

\section{Related Work}
\label{sec:related}

\textbf{Adjoint tidal inversion.} Estimating tidal parameters with adjoint methods has a long history in physical oceanography~\cite{bennett1992}. Global inverse models combine the linearized tidal equations with altimetry through generalized inversion~\cite{egbert1994,egbert2002}, and the same machinery has been used to map open-ocean dissipation~\cite{zaron2006}. Spatially varying bottom-friction coefficients have been estimated by variational assimilation in regional shelf models~\cite{das1991,lardner1993,heemink2002,lu2006}. More recently, automatically generated discrete adjoints have been used to calibrate spatially varying bottom friction against tide-gauge data at a cost independent of the number of parameters~\cite{karna2022}, and adjoint sensitivities have been used to design friction-estimation problems that avoid overfitting~\cite{warder2022}. \hct{} builds directly on this tradition and adds three elements: a plug-in interface that admits arbitrary differentiable closures, from 4-parameter laws to $10^4$-parameter networks, under one adjoint; architectural admissibility guarantees that hold for every parameter vector; and the reuse of the solve's by-products as a systematic audit of the fitted closure. To our knowledge, \hct{} is the first framework that trains tidal dissipation closures of arbitrary size, from physical laws to neural networks, through a round-off-exact frequency-domain global solve and audits them with the by-products of that same solve.

\textbf{Hard constraints and differentiable solvers.} Other hard-constraint approaches embed constraints in the network output~\cite{lu2021,beucler2021} or differentiate through time-stepping solvers~\cite{kochkov2021,kochkov2024,thuerey2021}. Physics has likewise been built into the architecture of compact networks for atmospheric turbulence~\cite{kriuk2026pstnet}, and hybrid physics--ML frameworks have been applied to permafrost-risk mapping~\cite{kriuk2026permafrost} and seismic inference~\cite{kriuk2026poseidon}. \hct{} targets a linear, frequency-domain operator, so an exact gradient costs one back-substitution and conservation is a discrete identity rather than a penalty.

\textbf{Spatial validation and atlas benchmarks.} Spatially blocked cross-validation is established in ecology and remote sensing~\cite{roberts2017,ploton2020,valavi2019}. \hct{} brings it to learned geophysical closures and combines it with physics-specific axes (frequency transfer, energy budgets and projection ensembles) that are available because the solver is in the loop. Comparisons with tide gauges and with altimetric atlases are the standard accuracy test for global tide models~\cite{stammer2014,ray2013}; we use one to separate the representativeness error of the grid from the error of the closure.

\section{Framework and Methodology}
\label{sec:method}

\subsection{Design Principles}
\label{sec:principles}
\hct{} follows five principles, each realized by a component in Fig.~\ref{fig:arch}:
\begin{itemize}
\item[\textbf{P1}] \emph{Solve, do not penalize.} The governing equations are satisfied to solver precision. Data enter only through the gauge operator.
\item[\textbf{P2}] \emph{Constrain the output, not the loss.} Positivity of drag is imposed by the parameterization, so every $\theta$ is admissible.
\item[\textbf{P3}] \emph{Learn only what is uncertain.} The closure predicts the drag field and one SAL scalar; wave dynamics, rotation and forcing are not learned.
\item[\textbf{P4}] \emph{Scale-free gradients.} As in classical adjoint inversion, gradient cost does not depend on the number of parameters; \hct{} extends this property to arbitrary learned closures, so a 4-parameter law and a $10^4$-parameter network are compared at equal computational budget.
\item[\textbf{P5}] \emph{Audit-ready by construction.} Energy budgets, dissipation maps and perturbed-geometry solves are differentiable by-products, not separate post-processing.
\end{itemize}

\subsection{Governing Equations and Discretization}
\label{sec:physics}
For each constituent $k$ with frequency $\omega_k$, fields are written as $\mathrm{Re}[\,\cdot\,e^{i\omega_k t}]$, and we solve the linearized, depth-averaged Laplace tidal equations
\begin{align}
i\omega_k\mathbf{u}+f\,\hat{\mathbf{z}}\times\mathbf{u}+\kappa_\theta\,\mathbf{u} &= -g\nabla\Phi, \label{eq:mom}\\
i\omega_k\eta+\nabla\cdot(H\mathbf{u}) &= 0, \label{eq:cont}
\end{align}
with $\Phi=(1-\beta_\theta)\eta-\eta^{\mathrm{eq}}_k$. Here $\mathbf{u}$ is the depth-averaged velocity, $\eta$ the elevation, $H$ the depth, $f=2\Omega\sin\varphi$ the Coriolis parameter and $\kappa_\theta$ the linear damping rate of Sec.~\ref{sec:closure}. The factor $\beta_\theta$ is a scalar SAL approximation~\cite{hendershott1972}. The equilibrium tide $\eta^{\mathrm{eq}}_k=\gamma_kA_kP_k(\varphi)e^{im_k\lambda}$, with $P_k=\cos^2(\varphi)$ (semidiurnal) or $\sin(2\varphi)$ (diurnal), includes the body-tide Love factor $\gamma_k=1+k_2-h_2$, which is $0.693$ for semidiurnal constituents and takes frequency-dependent values for diurnal constituents affected by the free core nutation. Complex amplitudes $Z=|Z|e^{-iG}$ follow the same convention as the gauge analysis, $\eta(t)=f_k\,\mathrm{Re}[Z\,e^{i(V_k+u_k)}]$, with nodal factors $f_k,u_k$ and astronomical argument $V_k$. In \eqref{eq:mom}, $i\omega_k\mathbf{u}$ is the acceleration of the flow at the constituent frequency; the Coriolis term $f\,\hat{\mathbf{z}}\times\mathbf{u}$ deflects the flow without doing work; $\kappa_\theta\,\mathbf{u}$ is the drag, which opposes the flow and removes energy; and $-g\nabla\Phi$ is the pressure-gradient force. Equation~\eqref{eq:cont} expresses mass conservation: the sea surface rises where the depth-integrated transport $H\mathbf{u}$ converges and falls where it diverges. Together, \eqref{eq:mom}--\eqref{eq:cont} describe a rotating, damped long gravity wave forced by the tidal potential.

\textbf{Forcing and body tide.} The potential $\Phi$ has two parts. The first, $(1-\beta_\theta)\eta$, is the sea-surface elevation reduced by self-attraction and loading. The second, $-\eta^{\mathrm{eq}}_k$, is the equilibrium tide, i.e., the elevation the ocean would take if it responded instantaneously to the tidal potential of the Moon and the Sun. Its gradient is the tide-generating force, which is the only forcing in the framework. The ocean cannot follow this potential instantaneously: it responds as a dynamic wave that lags the forcing and is amplified near basin resonances, which is why the equations must be solved. The solid Earth deforms under the same potential (the body tide). The seafloor rises by $h_2$ times the equilibrium tide, and the associated mass redistribution adds $k_2$ times the tidal potential. Because the ocean tide is measured relative to the moving seafloor, the net forcing is the equilibrium tide scaled by the body-tide Love factor $\gamma_k=1+k_2-h_2$.

\textbf{Physical assumptions.} Positivity of $\kappa_\theta$ is essential: if $\kappa_\theta<0$, the term $-\kappa_\theta\mathbf{u}$ would do positive work on the flow and inject energy that has no physical source. The depth-averaged (shallow-water) form is appropriate because tidal wavelengths ($10^3$--$10^4$\,km) greatly exceed the ocean depth, so the pressure is hydrostatic. Linearization neglects momentum advection, replaces quadratic bottom stress by a linear drag and assumes $|\eta|\ll H$. These assumptions hold well in the open ocean but degrade in shallow shelf seas, where nonlinear overtides and compound tides (e.g., $M_4$, $MS_4$) are generated; such constituents are excluded here. The loss of barotropic energy to internal tides is not resolved by a single-layer model and is represented through the drag closure.

The equations are discretized on an Arakawa C-grid~\cite{arakawa1977} on the sphere at $2^\circ$ (and at $1^\circ$ in Sec.~\ref{sec:res:res}), restricted to $|\varphi|\le80^\circ$ and $H>10$\,m. Velocities sit on cell faces and elevations at cell centres. No-normal-flow conditions are imposed on every face between a wet and a dry cell and at the latitude cut-off, i.e., these normal velocities are removed from the unknown vector, so no energy enters or leaves through the boundary. The discrete gradient and divergence are then negative adjoints of each other under area-weighted inner products (summation by parts) without boundary terms. Coriolis uses shared-corner weights that make it skew-symmetric in the energy norm, so it does no discrete work, and the discrete energy balance reduces exactly to $W(\theta)=D(\theta)$ (Sec.~\ref{sec:energy}). Because $\kappa_\theta\ge\kappa_{\min}>0$, $1-\beta_\theta>0.8$ and $\omega_k\neq0$, any null vector of $\mathbf{A}$ would have zero dissipation, hence zero velocity, hence (by \eqref{eq:cont}) zero elevation. $\mathbf{A}(\theta;\omega_k)$ is therefore non-singular for every $\theta$. Unknowns are ordered as [$u$ faces $|$ $v$ faces $|$ $\eta$ cells], which gives the sparse system $\mathbf{A}(\theta;\omega_k)\,\mathbf{x}_k=\mathbf{b}_k(\theta)$ of Sec.~\ref{sec:adjoint}.

\subsection{Closure Interface and Constraint Layer}
\label{sec:closure}
All closures share one interface, which is what makes \hct{} a framework rather than a single model. Given $\theta$ and frequency $\omega_k$, a closure returns a damping-rate field $\kappa_\theta(\mathbf{x};\omega_k)$ [s$^{-1}$] on wet cells and an SAL factor $\beta_\theta$. Any differentiable map with this signature can be plugged in without touching the solver, the adjoint or the audit.

\textbf{Features.} The learnable closures read a fixed, gauge-independent feature vector
\begin{equation}
\begin{aligned}
\bm\phi(\mathbf{x}) = \Big[&\tfrac12\ln\tfrac{H}{1\,\mathrm{km}},\;\ln\!\big(1+\tfrac{|\nabla H|}{0.005}\big),\\
&\sin\varphi,\;|\sin\varphi|,\;\min\!\big(\tfrac{|f|}{\omega},3\big)\Big],
\end{aligned}
\label{eq:features}
\end{equation}
and each model selects a subset of it. Unless a feature list is given in brackets, learnable closures use the three features $(\tfrac12\ln(H/1\,\mathrm{km}),\ \ln(1+|\nabla H|/0.005),\ \sin\varphi)$. The scalings are fixed constants, never estimated from data, which rules out leakage from test gauges through normalization statistics.

The first feature is a log-depth. It separates shelf seas, where bottom friction dominates, from the deep ocean, where conversion into internal tides dominates. The second is a compressed bathymetric slope, a proxy for internal-tide generation; because it is computed from bathymetry block-averaged to $2^\circ$, it smooths out small-scale seafloor roughness and therefore under-represents conversion. The terms $\sin\varphi$ and $|\sin\varphi|$ locate a cell in latitude and are proportional to $f$ and $|f|$: $\sin\varphi$ distinguishes the hemispheres, whereas $|\sin\varphi|$ is hemispherically symmetric. The last feature, $|f|/\omega$, decides whether freely propagating internal waves exist at frequency $\omega$, which requires $\omega>|f|$, i.e., equatorward of the critical latitude ($74.5^\circ$ for $M_2$); the cap at 3 keeps it bounded near the poles.

\textbf{Constraint layer.} The raw output $h_\theta(\mathbf{x})$ of a learnable closure is mapped to a drag velocity, a damping rate and an SAL factor:
\begin{align}
r_\theta(\mathbf{x}) &= r_{\mathrm{ref}}\exp\!\big(s\tanh(h_\theta(\mathbf{x})/s)\big), \label{eq:rmap}\\
\kappa_\theta(\mathbf{x}) &= \frac{r_\theta(\mathbf{x})}{\max(H,H_{\min})} + \kappa_{\min}, \label{eq:kappa}\\
\beta_\theta &= 0.2\,\sigma(\theta_\beta), \label{eq:beta}
\end{align}
with $r_{\mathrm{ref}}=10^{-3}$\,m\,s$^{-1}$, $s=6$, $\kappa_{\min}=10^{-6}$\,s$^{-1}$ and $H_{\min}=10$\,m. Hence $r_\theta\in[2.5\times10^{-6},\,0.40]$\,m\,s$^{-1}$, $\beta_\theta\in(0,0.2)$ and $\kappa_\theta\ge\kappa_{\min}>0$ \emph{for every $\theta$}.

\textbf{Physical interpretation.} The quantity $r_\theta$ in \eqref{eq:rmap} is a drag velocity: the bottom stress is $\bm\tau_b=\rho\,r_\theta\mathbf{u}$, so $r_\theta$ measures how strongly the seabed resists the flow, and the local dissipation rate $\rho\,r_\theta|\mathbf{u}|^2$ depends on both $r_\theta$ and the current speed. Two locations with similar $r_\theta$ can therefore dissipate very different amounts of energy. Physical bottom stress is quadratic, $\bm\tau_b=\rho c_d|\mathbf{u}|\mathbf{u}$; the linear form corresponds to $r\approx c_dU$ for a representative tidal speed $U$, so a spatially variable $r_\theta$ can absorb the stronger currents of shelf seas (large $r$) and the weak currents of the deep ocean (small $r$). Dividing by depth in \eqref{eq:kappa} converts a stress that acts at the bed into a damping rate for the depth-averaged velocity of the whole water column: a shallow column has little mass per unit area and is decelerated quickly, a deep column slowly. The ratio $\kappa_\theta/\omega$ sets the dynamical regime. For $\kappa_\theta\ll\omega$ (deep ocean) the tide propagates as a weakly damped wave; for $\kappa_\theta\sim\omega$ (shallow shelf seas) a substantial fraction of the wave energy is dissipated every cycle; for $\kappa_\theta\gg\omega$ the flow is friction-dominated and no longer wave-like. Because $\omega_k$ and $|f|$ are comparable at mid-latitudes, the inviscid version of \eqref{eq:mom}--\eqref{eq:cont} admits near-resonant responses and a nearly singular operator; the floor $\kappa_{\min}$ guarantees a small dissipation everywhere and hence G1 below. The SAL factor \eqref{eq:beta} dissipates no energy. It represents the gravitational self-attraction of the tidal bulge and the loading deformation of the seabed under its weight, which together reduce the effective restoring force to $(1-\beta_\theta)g$. Because the free gravity-wave speed is $c=\sqrt{(1-\beta_\theta)gH}$, small changes in $\beta_\theta$ shift basin resonances and can change tidal amplitudes appreciably. A single scalar $\beta_\theta$ captures only the spatially uniform, in-phase part of self-attraction and loading. The full effect is a nonlocal convolution of $\eta$ with a kernel built from the load Love numbers, so its ratio to the local elevation varies regionally and carries a phase lag. Because $\beta_\theta$ is fitted here, it can also absorb other model errors and should be read as an effective parameter rather than a measured constant.

Together with the discrete structure of Sec.~\ref{sec:physics}, this yields three architectural guarantees:
\begin{itemize}
\item[\textbf{G1}] the operator $\mathbf{A}(\theta)$ is strictly dissipative and therefore non-singular;
\item[\textbf{G2}] discrete work input equals discrete dissipation, $W(\theta)=D(\theta)$, to round-off;
\item[\textbf{G3}] for the learnable closures, drag is bounded above and below, so frictionless or fully blocked solutions cannot occur.
\end{itemize}
G1 and G2 hold for every closure, including \phys{} below, whose terms are positive through its log-parameterization; G3 is specific to \eqref{eq:rmap}. G2 follows from the summation-by-parts construction and therefore serves as an exact run-time integrity check of the assembled operator, the forcing and the diagnostics.

\textbf{Closure families.} \emph{(i)~\phys{}} is a 4-parameter physical law with background drag, a shelf enhancement and a topographic internal-wave term in the spirit of~\cite{jayne2001}:
\begin{equation}
\begin{aligned}
\kappa ={}& \frac{r_0}{H} + \frac{r_{\mathrm{sh}}}{H}e^{-H/200\,\mathrm{m}}\\
&+ c_{\mathrm{it}}|\nabla H|^2\,\sigma\!\Big(\frac{H-500\,\mathrm{m}}{100\,\mathrm{m}}\Big)\chi_\omega + \kappa_{\min},
\end{aligned}
\label{eq:phys}
\end{equation}
with $\theta=(\ln r_0,\ln c_{\mathrm{it}},\mathrm{logit}(\beta/0.2),\ln r_{\mathrm{sh}})$. For \phys{}, $\chi_\omega\equiv1$. For \physfc{}, $\chi_\omega=\sqrt{(1-f^2/\omega^2)_+}$, which switches off internal-wave drag poleward of the critical latitude. Each term of \eqref{eq:phys} is non-negative, so the four sinks (mechanism-agnostic background bottom friction $r_0/H$, shelf enhancement concentrated in $H\lesssim200$\,m, topographic internal-wave drag active in water deeper than $\approx500$\,m, and the floor $\kappa_{\min}$) add as independent dissipation rates, and none of them can inject energy. Each term enters \eqref{eq:mom} as a damping rate that removes energy from the flow at a rate proportional to $\kappa|\mathbf{u}|^2$, which is why their sum can only dissipate. The background stress associated with $r_0$ acts on the whole water column, so the same drag velocity gives a high damping rate in shallow water and a weak one in the deep ocean. On its own, $r_0$ underestimates shelf damping; $r_{\mathrm{sh}}$ is therefore a drag-velocity coefficient that supplies additional damping in shallow water. The coefficient $c_{\mathrm{it}}$ sets the strength of the barotropic energy loss to internal-tide generation and is confined to the deep ocean by the sigmoid in \eqref{eq:phys}. Because $r_{\mathrm{sh}}$ and $c_{\mathrm{it}}$ act in different depth ranges, double counting of dissipation is reduced, although not eliminated. Richer conversion schemes~\cite{green2013,buijsman2015} plug into the same interface. \emph{(ii)~\poly{d}}: $h_\theta(\bm\phi)=\sum_{|\alpha|\le d}c_\alpha\bm\phi^\alpha$, composed with \eqref{eq:rmap}. \emph{(iii)~MLP}: $h_\theta=\mathbf{w}_{o}^{\!\top}\tanh(\mathbf{W}_2\tanh(\mathbf{W}_1\bm\phi+\mathbf{b}_1)+\mathbf{b}_2)+b_o$, with width $w\in\{8,32,128\}$ and depth 2, i.e., 114, 1\,218 and 17\,154 parameters with three features (including $\theta_\beta$). The output layer is initialized at $10^{-3}$ scale, so every network starts from a near-uniform drag $r\approx r_{\mathrm{ref}}$.

\textbf{Regularizers.} \phys{} uses a Gaussian prior $\Omega=\|\theta-\theta_0\|^2$ ($\lambda=10^{-3}$). The MLPs use weight decay on weight matrices only ($\lambda\in\{10^{-6},10^{-4},10^{-3}\}$), and \poly{d} penalizes its non-constant coefficients.

\subsection{Implicit Adjoint Layer}
\label{sec:adjoint}
For constituent $k$, the state $\mathbf{x}_k=(\mathbf{u},\mathbf{v},\bm\eta)_k\in\mathbb{C}^{n}$ ($n=28\,383$ at $2^\circ$, $114\,524$ at $1^\circ$) solves
\begin{equation}
\mathbf{r}_k(\mathbf{x};\theta) \equiv \mathbf{A}(\theta;\omega_k)\,\mathbf{x}-\mathbf{b}_k(\theta) = \mathbf{0}.
\end{equation}
By the implicit function theorem, $\partial\mathbf{x}_k/\partial\theta = -\mathbf{A}_k^{-1}\partial_\theta\mathbf{r}_k$, so the reverse-mode vector--Jacobian product is
\begin{equation}
\begin{aligned}
\nabla_\theta\mathcal{L} &= \partial_\theta\mathcal{L}\big|_{\mathbf{x}} -\sum_{k\in\mathcal{K}}\mathrm{Re}\Big[\big(\partial_\theta\mathbf{r}_k\big)^{H}\bm\lambda_k\Big],\\
\mathbf{A}_k^{H}\bm\lambda_k &= \partial_{\mathbf{x}_k}\mathcal{L},
\end{aligned}
\label{eq:adjoint}
\end{equation}
where $\partial_{\mathbf{x}_k}\mathcal{L}$ is the conjugate (Wirtinger) gradient of the real loss with respect to the complex state, and $\partial_\theta\mathcal{L}|_{\mathbf{x}}$ collects the explicit dependence of the regularizer and of the energy penalty (through $\kappa_\theta$) on $\theta$. This is the adjoint-state method~\cite{plessix2006,errico1997}. It requires one back-substitution per constituent with the \emph{already computed} sparse LU factors~\cite{li2005}. We implement it as a custom autograd primitive~\cite{paszke2019} in three steps:
\begin{enumerate}
\item Compute $\bar{\mathbf{x}}_k=\mathbf{A}_k^{-1}\mathbf{b}_k$ outside the graph.
\item Evaluate the residual $\mathbf{r}_k(\bar{\mathbf{x}}_k;\theta)$ \emph{inside} the graph with a matrix-free routine that matches the assembled $\mathbf{A}$ exactly.
\item Return $\mathbf{x}_k=\bar{\mathbf{x}}_k-\mathcal{S}(\mathbf{r}_k)$, where $\mathcal{S}(\mathbf{r})=\mathbf{A}_k^{-1}\mathbf{r}$ in the forward pass and $\mathcal{S}^{*}(\mathbf{g})=\mathbf{A}_k^{-H}\mathbf{g}$ in the backward pass.
\end{enumerate}
The forward value equals $\bar{\mathbf{x}}_k$ up to a residual-sized correction, and the gradient is exactly \eqref{eq:adjoint}. Automatic differentiation carries $\partial_\theta\mathbf{r}_k$ through \eqref{eq:rmap}--\eqref{eq:beta} into any closure. Closures are therefore plug-ins, and extra parameters only add the cost of evaluating the closure on $\sim\!10^4$ cells.

\textbf{Solver abstraction.} The primitive needs only two operations: a solve with $\mathbf{A}_k$ and a solve with $\mathbf{A}_k^H$. At $2^\circ$ and at $1^\circ$ both are served by one sparse LU factorization, and G2 holds to round-off on both grids (Sec.~\ref{sec:res:res}). At still finer resolution the same primitive accepts a preconditioned Krylov solver for the forward and adjoint systems without any change to the closure interface, the constraint layer or the audit. G1 and G3 are properties of the operator and the parameterization and hold at any resolution; with an iterative solver G2 holds to the solver tolerance, and $|W-D|/W$ becomes a direct monitor of that tolerance.

\subsection{Differentiable Energy Diagnostics}
\label{sec:energy}
From the solved fields we compute tidal work and dissipation,
\begin{align}
W(\theta) &= \tfrac12\rho g\textstyle\sum_{c}A_c\,\mathrm{Re}\big[\Phi_c\,(-i\omega\eta_c)^{*}\big], \label{eq:work}\\
D(\theta) &= \tfrac12\rho\textstyle\sum_{f}A_f\,\kappa_f H_f|u_f|^2, \label{eq:diss}
\end{align}
where $\Phi$ is defined in Sec.~\ref{sec:physics}, $c$ runs over cells and $f$ over velocity faces, and $\kappa_f$, $H_f$ are the damping rate and depth interpolated to face $f$; by \eqref{eq:cont}, $-i\omega\eta=\nabla\cdot(H\mathbf{u})$, so \eqref{eq:work} is the time-mean rate of work done by the forcing. The shelf share is $f_{\mathrm{sh}}=D_{H<500\,\mathrm{m}}/D$. By G2, $W\equiv D$, so $|W-D|/W$ serves as a run-time integrity check.

\textbf{Physical interpretation.} The sum in \eqref{eq:work} runs over all cells. In each cell, $-i\omega\eta_c$ is minus the rate of change of elevation, which mass conservation equates to the divergence of the volume transport. The real part and the factor $\tfrac12$ come from averaging over a tidal cycle: the time mean of the product of two sinusoids with complex amplitudes $a$ and $b$ is $\tfrac12\mathrm{Re}[ab^{*}]$. The elevation part of $\Phi_c$, $(1-\beta_\theta)\eta_c$, does no net work over a cycle, because the energy spent raising the water surface is returned when the level falls; the net work is therefore done by the equilibrium-tide term alone. $D$ is the power removed by drag. The drag force per unit mass, $-\kappa\mathbf{u}$, acts on a water column of mass $\rho H$ per unit area, and time averaging again gives the factor $\tfrac12$. For the learnable closures, $\kappa H=r_\theta+\kappa_{\min}H$, so the depth cancels and the energy removed per unit area is $\tfrac12\rho\,(r_\theta+\kappa_{\min}H)|\mathbf{u}|^2$. This loss depends on the drag velocity and on the square of the current speed, so the same $r_\theta$ removes far more energy where the currents are strong, which is why dissipation is higher on the shelves. Averaged over a tidal cycle, the ocean stores no net energy, so $W(\theta)=D(\theta)$. The forcing does net work only because the ocean tide lags the equilibrium tide, and this lag is produced by dissipation.

Both quantities are differentiable through the implicit layer. This enables the optional admissibility penalty
\begin{equation}
\mathcal{E}(\theta)=\mu_D\big(D_{\Mtwo}/D^{\star}-1\big)^2+\mu_{\mathrm{sh}}\big(f_{\mathrm{sh}}-\tfrac23\big)^2,
\label{eq:energypen}
\end{equation}
with $D^{\star}=2.4$\,TW and both targets taken from altimetric estimates~\cite{egbert2000,egbert2001}, which place about two-thirds of $\Mtwo$ dissipation in shallow seas. The sweep of Sec.~\ref{sec:res:energy} sets $\mu_D=\mu_{\mathrm{sh}}=\mu$; an ablation separates the global-budget term ($\mu_D$) from the shelf-share term ($\mu_{\mathrm{sh}}$). At $2^\circ$ and $1^\circ$ the shelf term acts as a \emph{directional} target that steers dissipation toward the shelf cells the grid resolves; it becomes a quantitative constraint only on grids that resolve the shelf seas (at $1^\circ$ the unconstrained shelf share is still 19--25\%, Sec.~\ref{sec:res:res}). The penalty reuses the existing solve and adjoint, so it needs no additional linear solves.

\subsection{Objective and Optimization}
\label{sec:objective}
Let $\mathcal{T}$ denote the training gauges, $\hat Z_{gk}$ the observed complex harmonic constants and $\mathcal{K}$ the fitted constituents (default $\{M_2,S_2,K_1,O_1\}$). We minimize
\begin{equation}
\begin{aligned}
\mathcal{L}(\theta)={}&\sum_{k\in\mathcal{K}}\frac{\sum_{g\in\mathcal{T}}w_g\big|\eta_k(\mathbf{x}_g;\theta)-\hat Z_{gk}\big|^2}{\sum_{g\in\mathcal{T}}w_g}\\
&+\lambda\,\Omega(\theta)+\mathcal{E}(\theta),
\end{aligned}
\label{eq:loss}
\end{equation}
with density weights $w_g\propto1/|\{g':d(g,g')<500\,\mathrm{km}\}|$ (normalized to mean 1) that keep dense coastal clusters from dominating the fit. We use L-BFGS~\cite{liu1989} (history 10, strong-Wolfe line search, at most 60 iterations and PyTorch's default budget of $\approx$75 objective evaluations) and keep the best iterate (Algorithm~\ref{alg:train}). The fixed evaluation budget gives every closure the same compute and acts as an implicit early-stopping regularizer for the high-capacity closures.

\begin{algorithm}[t]
\caption{\hct{} training step (one $\mathcal{L}$, $\nabla_\theta\mathcal{L}$ evaluation)}
\label{alg:train}
\begin{algorithmic}[1]
\Require closure $g_\theta$, features $\bm\phi$, gauges $(\mathbf{x}_g,\hat Z_{gk},w_g)$, constituents $\mathcal{K}$
\State $\mathcal{L}\gets0$
\For{$k\in\mathcal{K}$}
  \State $(\kappa,\beta)\gets$ constraint layer$\big(g_\theta(\bm\phi;\omega_k)\big)$
  \State assemble $\mathbf{A}_k(\kappa,\beta)$; \ $\mathbf{L}_k\mathbf{U}_k\gets\mathrm{LU}(\mathbf{A}_k)$
  \State $\bar{\mathbf{x}}_k\gets\mathbf{U}_k^{-1}\mathbf{L}_k^{-1}\mathbf{b}_k$ \Comment{outside graph}
  \State $\mathbf{r}_k\gets\mathbf{A}_k(\theta)\bar{\mathbf{x}}_k-\mathbf{b}_k(\theta)$ \Comment{inside graph}
  \State $\mathbf{x}_k\gets\bar{\mathbf{x}}_k-\mathcal{S}_k(\mathbf{r}_k)$ \Comment{bwd: $\mathbf{A}_k^{-H}$}
  \State $\mathcal{L}\mathrel{+}=\sum_{g}w_g|\mathcal{P}_g\mathbf{x}_k-\hat Z_{gk}|^2/\sum_g w_g$
  \If{energy penalty \textbf{and} $k=\Mtwo$}
    \State $\mathcal{L}\mathrel{+}=\mathcal{E}(\mathbf{x}_k,\kappa)$ \Comment{Eq.~\eqref{eq:energypen}}
  \EndIf
\EndFor
\State $\mathcal{L}\mathrel{+}=\lambda\Omega(\theta)$; backpropagate; return $(\mathcal{L},\nabla_\theta\mathcal{L})$
\end{algorithmic}
\end{algorithm}

\section{Experimental Setup and Audit Protocol}
\label{sec:setup}

\textbf{Data.} Bathymetry is ETOPO1~\cite{amante2009}, block-averaged to $2^\circ$: $90\times180$ cells, 9\,757 of them wet, giving 28\,383 unknowns. Tide-gauge records are UHSLC hourly data for 2018~\cite{caldwell2015}. Each station is snapped to the nearest wet cell in the $3\times3$ neighbourhood of the cell containing it (the gauge operator $\mathcal{P}_g$ selects that cell's elevation), and kept if it has $\ge60\%$ hourly coverage and $\ge300$ days of span. Kept records undergo least-squares harmonic analysis for 20 constituents with nodal corrections~\cite{pugh2014}, and must then have residual s.d.\ $\le0.5$\,m and $|M_2|\le6$\,m. This yields 230 gauges. A fixed exclusion list removes four (UHSLC IDs 579, 600, 684 and 833), leaving 226 evaluated gauges.
For the refinement test (E9) the same bathymetry is block-averaged to $1^\circ$ (38\,927 wet cells, 114\,524 unknowns), and the same 226 gauges are evaluated with the same fold assignment.

\textbf{Scoring at coastal gauges.} Most UHSLC stations are coastal, so part of every model's error at $2^\circ$ is sub-grid coastal response that no basin-scale closure represents. This component is common to all closures and baselines, because all are scored on the same gauges against the same grid. The paired, gauge-level comparisons below therefore compare closures under identical representativeness conditions, and the framework's per-gauge outputs make every comparison traceable. We quantify this component explicitly with the atlas benchmark below and report every score also for two strata: \emph{open-ocean} gauges, whose snapped cell is at least 1\,000\,m deep (131 gauges), and \emph{coastal} gauges (95).

\textbf{External benchmark (E8).} We use the altimetric GOT4.10c atlas~\cite{ray2013} ($0.5^\circ$ grid, ocean-tide elevations of $M_2$, $S_2$, $K_1$ and $O_1$) as an external reference in two ways. At the gauges, the atlas is evaluated (i)~at the gauge location on its native grid, which measures how well a mature altimetric solution reproduces the gauge constants, and (ii)~at the same $2^\circ$ cell that the gauge operator $\mathcal{P}_g$ samples. The second score is a \emph{representativeness floor}: the error that a near-perfect tidal field incurs merely because it is represented at $2^\circ$. Atlas values are valid at 221 of the 226 gauges (128 open, 93 coastal), and all benchmark comparisons use this subset. Over the global field, we compare each fitted solution with the atlas averaged into the $2^\circ$ cells, separately over deep ($H\ge1\,000$\,m) and shallow wet cells.
GOT4.10c is a residual analysis of altimetry relative to prior hydrodynamic solutions and inherits part of their structure in shallow water, so we treat it as a reference rather than as truth. Moreover, because a closure fitted to gauges can partly compensate sub-grid response, the floor is a reference level rather than a strict lower bound.

\textbf{Resolution test (E9).} To check whether a learned gain depends on the grid, we repeat the hash-CV fits of \phys{}, \nn{8} and \nn{32}[$H$, slope, $|\sin\varphi|$] at $1^\circ$ with unchanged hyperparameters, folds and evaluation budget, and refit the same three closures at $2^\circ$ with identical solver settings so that the comparison is paired. Both grids use sparse LU with a symmetric fill-reducing ordering, which at $1^\circ$ factorizes the 114\,524-unknown system in 0.17\,s. The representativeness floor is recomputed on both grids by the E9 script for the evaluated gauges: 22.6\,cm at $2^\circ$ (22.8\,cm on the 221-gauge benchmark subset of E8) and 18.0\,cm at $1^\circ$.

\textbf{Baselines.} (i)~A complex GP with an exponential kernel on great-circle distance; the length scale (300--5\,000\,km) and nugget are chosen per constituent by leave-one-out. (ii)~The nearest training gauge. (iii)~The equilibrium tide. All baselines use only the training gauges of the same fold. The GP is the strongest of the three and serves as the data-only reference below.

\textbf{Metrics.} $\mathrm{RMS}_k=100\sqrt{\sum_g w_g|Z_{gk}-\hat Z_{gk}|^2/(2\sum_g w_g)}$\,cm is the time-domain RMS of the complex error. Skill is the root-sum-square (RSS) over constituents. We report fold means\,$\pm$\,s.d.\ and the \emph{pooled} RMS over all 226 gauges, each predicted exactly once. Paired differences use a gauge-level bootstrap with 4\,000 resamples and 95\% CIs, complemented by per-fold and per-block win counts that resolve each comparison region by region. Strata scores and the excess of each model over the representativeness floor are computed from the same held-out, gauge-level predictions with the same bootstrap.

\textbf{Validation designs (A1).} \emph{Hash CV} assigns gauge $g$ to fold $\mathrm{SHA256}(\mathrm{seed}{:}\mathrm{id})\bmod5$, which is random in space. \emph{Block CV} partitions the gauges into five contiguous regions by spherical $k$-means with 25 restarts (Fig.~\ref{fig:blocksgain}a). Both designs cover the same 226 gauges, so pooled scores are directly comparable.

\textbf{Experiments and reproducibility.} Table~\ref{tab:exps} lists the suite. All 170 fits took 100\,min of L-BFGS time on one laptop CPU (Apple M4 Pro, 10 threads; Python 3.11, PyTorch 2.14, SciPy 1.17). The 30 fits of E9 are run by a separate script with ten worker processes and take 3.0\,min of wall-clock time. Each fit is cached by a content hash of its configuration, so runs shared between experiments are computed once. A run manifest records the software environment, the code commit, a digest of the gauge set and the exclusion list.

\begin{table}[t]
\centering\footnotesize
\caption{Experiment suite and audit axes. Fit counts are distinct fits (170 in total for E1--E5b; E9 adds 30 with its own solver settings); E3 ($\lambda=10^{-6}$), E5 ($\mu=0$), E5b (both terms), E6, E7 and E8 reuse earlier fits, which are not counted again.}
\label{tab:exps}
\setlength{\tabcolsep}{3pt}
\begin{tabular}{@{}llp{4.1cm}r@{}}
\toprule
ID & Axis & Protocol & Fits \\
\midrule
E0 & -- & energy identity; adjoint vs.\ central FD, 5 closures & 20 checks\\
E1 & skill & 5-fold hash CV, capacity ladder + feature ablation & 35\\
E2 & A1 & 5-fold spherical-block CV, 6 closures & 30\\
E3 & skill & weight decay $\lambda\times$ width & 30\\
E4 & A2 & fit $M_2$ only, score $M_2,S_2,K_1,O_1$ & 25\\
E5 & A3 & energy penalty $\mu\in\{0,.03,.1,.3,1\}$ & 40\\
E5b & A3 & term ablation: global budget only ($\mu_D{=}1,\mu_{\mathrm{sh}}{=}0$) & 10\\
E6 & A4 & distil \nn{8} into degree-1/2/3 laws & 15 evals\\
E7 & A5 & $+1$\,m sea level, 5-fold ensembles & 10 pairs\\
E8 & -- & GOT4.10c floor, open/coastal strata, field & 0\\
E9 & -- & $1^\circ$ vs.\ $2^\circ$: hash CV, 3 closures, atlas floor on both grids & 30\\
\bottomrule
\end{tabular}
\end{table}

\section{Results}
\label{sec:results}

\subsection{Verification of the Framework (E0)}
\label{sec:res:verif}

\begin{figure}[t]
\centering
\begin{tikzpicture}
\begin{axis}[hct, width=\columnwidth, height=3.8cm,
  ymode=log, ymin=4e-9, ymax=3e-6, log origin=infty,
  xmin=0.2, xmax=20.8,
  xtick={2.5,6.5,10.5,14.5,18.5},
  xticklabels={\phys{},\physfc{},\poly{2},\nn{8},{\nn{8}+$\mathcal{E}$}},
  ylabel={rel.\ error (adjoint vs.\ FD)}, ymajorgrids,
  title={(a) gradient verification}, title style={font=\scriptsize, at={(0.02,0.97)}, anchor=north west}]
\addplot[ybar, bar width=5pt, bar shift=0pt, fill=cphys, draw=none] coordinates {(1,6.2e-8)(2,7.2e-8)(3,7.8e-7)(4,8.3e-9)};
\addplot[ybar, bar width=5pt, bar shift=0pt, fill=csky, draw=none] coordinates {(5,6.5e-8)(6,5.4e-8)(7,7.1e-7)(8,9.7e-9)};
\addplot[ybar, bar width=5pt, bar shift=0pt, fill=cpoly, draw=none] coordinates {(9,4.2e-7)(10,2.5e-8)(11,2.0e-7)(12,1.5e-7)};
\addplot[ybar, bar width=5pt, bar shift=0pt, fill=cnn, draw=none] coordinates {(13,1.5e-8)(14,1.5e-7)(15,1.4e-7)(16,3.6e-8)};
\addplot[ybar, bar width=5pt, bar shift=0pt, fill=cnnd, draw=none] coordinates {(17,1.5e-8)(18,1.2e-7)(19,1.4e-7)(20,4.9e-8)};
\addplot[densely dashed, black, thin] coordinates {(0.2,1e-6)(20.8,1e-6)};
\end{axis}
\end{tikzpicture}

\vspace{1mm}
\begin{tikzpicture}
\begin{axis}[hct, width=\columnwidth, height=3.6cm, xmode=log,
  xmin=2, xmax=6e4, ymin=0, ymax=1.0,
  xlabel={trainable parameters $\dim\theta$},
  ylabel={s per $(\mathcal{L},\nabla\mathcal{L})$},
  ytick={0,0.25,0.5,0.75,1.0}, grid=major,
  title={(b) gradient cost}, title style={font=\scriptsize, at={(0.02,0.97)}, anchor=north west}]
\addplot[densely dashed, gray] coordinates {(2,0.65)(6e4,0.65)};
\addplot[only marks, mark=*, mark size=2.2pt, color=cphys] coordinates {(4,0.65)};
\addplot[only marks, mark=square*, mark size=2.2pt, color=cpoly] coordinates {(11,0.65)};
\addplot[only marks, mark=triangle*, mark size=2.8pt, color=cnn] coordinates {(114,0.66)(1218,0.68)(17154,0.75)};
\node[font=\tiny, anchor=south] at (axis cs:4,0.69) {\phys};
\node[font=\tiny, anchor=north] at (axis cs:11,0.61) {\poly{2}};
\node[font=\tiny, anchor=south] at (axis cs:114,0.70) {\nn{8}};
\node[font=\tiny, anchor=south] at (axis cs:1218,0.72) {\nn{32}};
\node[font=\tiny, anchor=south] at (axis cs:17154,0.79) {\nn{128}};
\node[font=\scriptsize, align=left, anchor=west] at (axis cs:3,0.22) {$+15\%$ cost for $4\,300\times$ parameters};
\end{axis}
\end{tikzpicture}
\caption{Framework verification. (a)~Relative error of implicit-adjoint vs.\ central finite-difference derivatives ($h=10^{-3}$), four per closure: the four parameter derivatives for \phys{} and \physfc{}, four random unit directions for the others; the dashed line marks $10^{-6}$. (b)~Wall-clock time per loss-and-gradient evaluation (four constituents), averaged over the E1 fits.}
\label{fig:verif}
\end{figure}
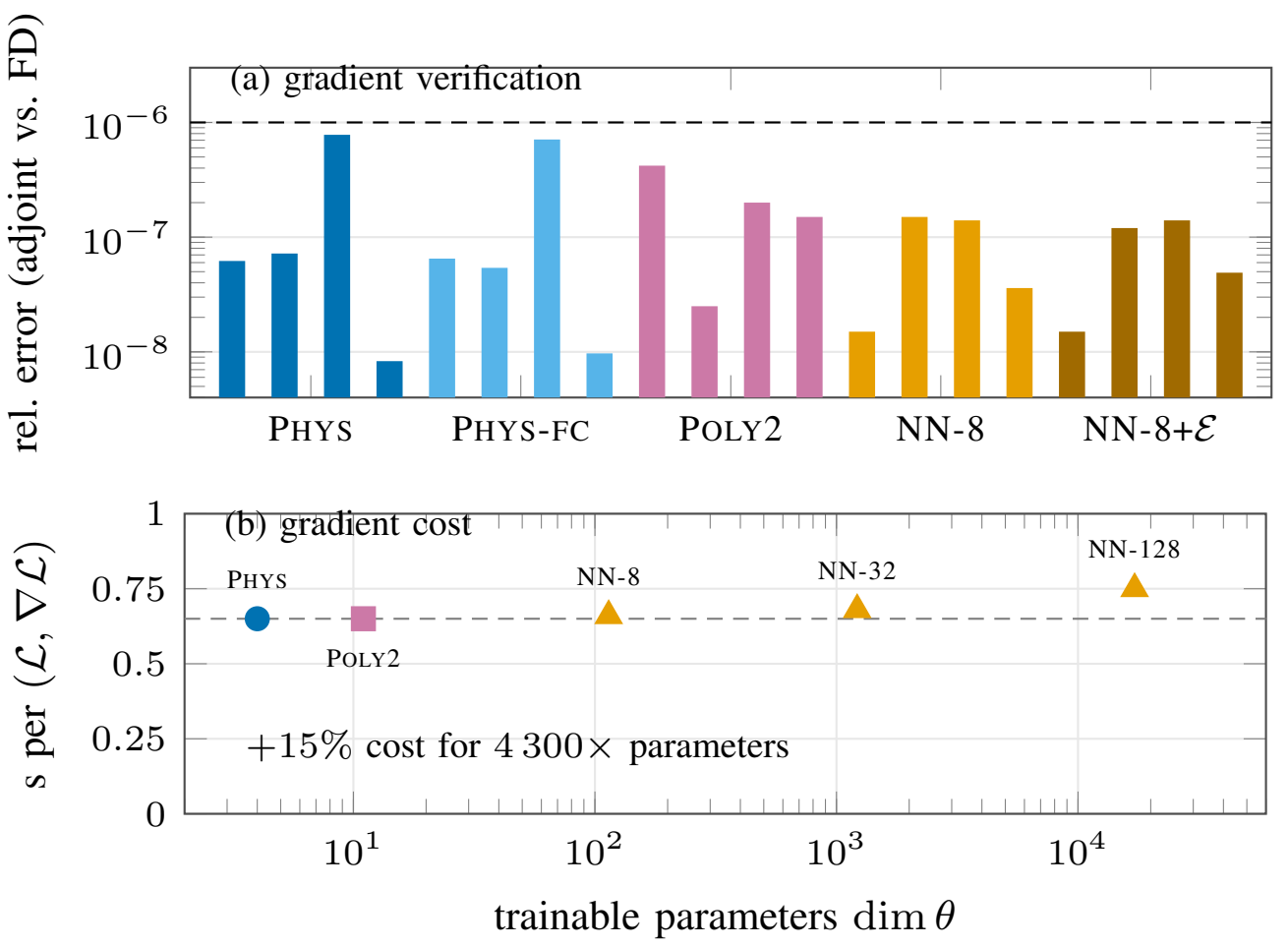

For all closures and both constituent species, the relative mismatch between discrete work and dissipation is $\le6.1\times10^{-16}$ (Table~\ref{tab:check}). The energy identity is therefore an exact property of the discrete architecture, not something the model learns. Across 20 derivative checks, which include the MLP with the energy penalty \eqref{eq:energypen} active, adjoint gradients agree with central finite differences to between $8.3\times10^{-9}$ and $7.8\times10^{-7}$ (Fig.~\ref{fig:verif}a), consistent with the $\mathcal{O}(h^2)$ truncation of the finite-difference reference. The same checks pass on an independent synthetic bathymetry with a different mesh (13\,656 wet cells, 40\,714 unknowns): the energy mismatch is $\le3.1\times10^{-16}$, the sparse-solve residual $\|\mathbf{A}\mathbf{x}-\mathbf{b}\|/\|\mathbf{b}\|\le8.4\times10^{-13}$, and the adjoint agrees with finite differences to $\le3.0\times10^{-6}$, with the largest deviation for the SAL parameter. The time per loss-and-gradient evaluation for four constituents rises from 0.65\,s for \phys{} to 0.75\,s for the 17\,154-parameter MLP: $+15\%$ for a $4\,300\times$ increase in $\dim\theta$ (Fig.~\ref{fig:verif}b). A central finite-difference gradient of the same MLP would cost $3.4\times10^4$ objective evaluations. The identity is not tied to the $2^\circ$ grid or to initial parameters: over all fitted models of the refinement test (Sec.~\ref{sec:res:res}), $|W-D|/W\le4.9\times10^{-16}$ at $2^\circ$ and $\le3.3\times10^{-16}$ at $1^\circ$, with solve residuals $\le4.5\times10^{-13}$ and $\le2.0\times10^{-12}$.

\begin{table}[t]
\centering\footnotesize
\caption{Verification at initial parameters (randomly perturbed for closures with more than six parameters): energy identity and adjoint gradient check.}
\label{tab:check}
\setlength{\tabcolsep}{4pt}
\begin{adjustbox}{max width=\columnwidth}
\begin{tabular}{@{}lrccc@{}}
\toprule
Closure & \#p & $|W{-}D|/W$, $M_2$ & $|W{-}D|/W$, $K_1$ & max grad err.\\
\midrule
\phys{}         & 4   & $2.1\times10^{-16}$ & $1.1\times10^{-16}$ & $7.8\times10^{-7}$\\
\physfc{}       & 4   & $2.1\times10^{-16}$ & $2.3\times10^{-16}$ & $7.1\times10^{-7}$\\
\poly{2}        & 11  & $1.8\times10^{-16}$ & $6.1\times10^{-16}$ & $4.2\times10^{-7}$\\
\nn{8}          & 114 & $1.9\times10^{-16}$ & $2.2\times10^{-16}$ & $1.5\times10^{-7}$\\
\nn{8} + energy & 114 & $1.9\times10^{-16}$ & $2.2\times10^{-16}$ & $1.4\times10^{-7}$\\
\bottomrule
\end{tabular}
\end{adjustbox}
\end{table}

\subsection{Physics-Constrained Skill and the Capacity Ladder (E1)}
\label{sec:res:hash}

\begin{figure}[t]
\centering
\begin{tikzpicture}
\begin{axis}[hct, width=\columnwidth, height=5.0cm, xmode=log,
  xmin=2, xmax=6e4, ymin=26, ymax=57,
  xlabel={trainable parameters $\dim\theta$}, ylabel={RMS [cm]},
  grid=major,
  legend columns=2, legend style={at={(0.5,1.02)}, anchor=south}]
\addplot[cnn, thick, mark=square, mark options={solid, fill=white}] coordinates {(114,32.7)(1218,30.6)(17154,29.2)};
\addlegendentry{in-sample (hash)}
\addplot[cnnd, thick, mark=*] coordinates {(114,34.4)(1218,35.3)(17154,35.2)};
\addlegendentry{pooled held-out (hash)}
\addplot[cnn, thick, densely dashed, mark=square, mark options={solid, fill=white}] coordinates {(114,32.4)(1218,30.7)(17154,29.6)};
\addlegendentry{in-sample (block)}
\addplot[cred, thick, densely dashed, mark=*, mark options={solid}] coordinates {(114,43.2)(1218,44.6)(17154,41.2)};
\addlegendentry{pooled held-out (block)}
\addplot[forget plot, only marks, mark=square, mark size=2.2pt, cphys, thick] coordinates {(4,37.2)};
\addplot[forget plot, only marks, mark=*, mark size=2.2pt, cphys] coordinates {(4,37.7)};
\addplot[forget plot, only marks, mark=*, mark size=2.2pt, cred] coordinates {(4,39.0)};
\addplot[forget plot, only marks, mark=square, mark size=2.2pt, cpoly, thick] coordinates {(11,35.5)};
\addplot[forget plot, only marks, mark=*, mark size=2.2pt, cpoly] coordinates {(11,39.4)};
\addplot[forget plot, gray, thin] coordinates {(2,47.6)(6e4,47.6)};
\addplot[forget plot, gray, thin, densely dashed] coordinates {(2,54.6)(6e4,54.6)};
\node[font=\tiny, gray!50!black, anchor=south west] at (axis cs:2.2,47.6) {GP (hash)};
\node[font=\tiny, gray!50!black, anchor=south west] at (axis cs:2.2,54.6) {GP (block)};
\node[font=\tiny, cphys, anchor=north] at (axis cs:4,36.8) {\phys};
\node[font=\tiny, cpoly, anchor=north] at (axis cs:11,35.0) {\poly{2}};
\draw[<->, cred!70, thin] (axis cs:30000,29.6) -- node[right, font=\tiny, align=left]{$+11.6$\\(pooled$-$in)} (axis cs:30000,41.2);
\end{axis}
\end{tikzpicture}
\caption{Capacity ladder. In-sample error (open squares) falls monotonically with width. Pooled held-out error (filled) stays flat under hash CV (solid) and is 6.0--9.3\,cm higher under block CV (dashed). \phys{} (blue/red dots at 4 parameters) is the most stable model across the two designs.}
\label{fig:capacity}
\end{figure}
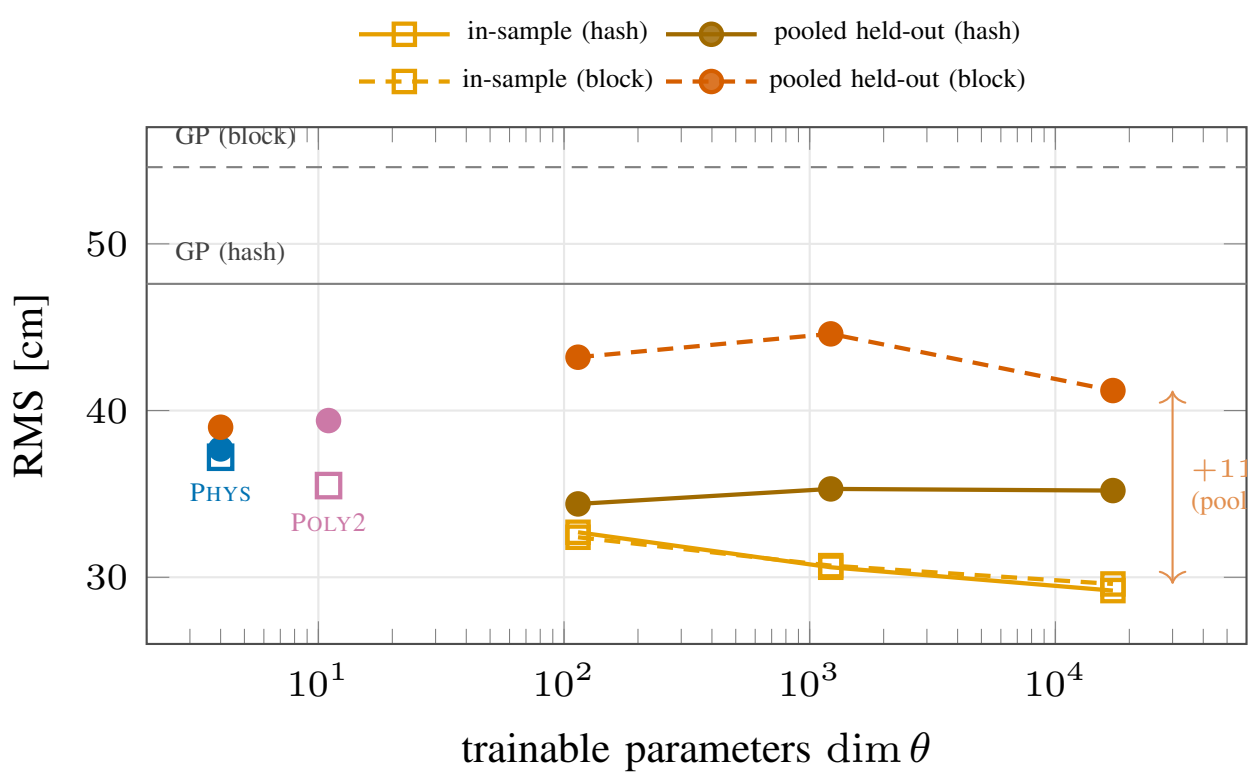

Table~\ref{tab:cv} summarizes hash CV. \textbf{All} physics-constrained closures beat every data-only baseline by a wide margin. \phys{} lowers the pooled GP error from 47.6 to 37.7\,cm ($-21\%$), and the best MLP reaches 33.9\,cm ($-29\%$). For every closure, $P(\text{model}\ge\text{GP})<2.5\times10^{-4}$. Under this design the neural closures improve significantly on \phys{}: \nn{8} by $-3.3$\,cm [$-5.7,-1.0$], winning 4/5 folds, and \nn{32}[$H$,slope,$|\sin\varphi|$] by $-3.8$\,cm [$-6.3,-1.6$], winning 5/5. The capacity ladder (Fig.~\ref{fig:capacity}) shows that the framework extracts this gain already at the smallest width. In-sample error falls monotonically with width (32.7$\to$30.6$\to$29.2\,cm) while held-out error stays flat (34.5, 35.4, 35.1\,cm). The generalization gap grows from $+1.9$ to $+6.0$\,cm, so the compact 114-parameter network is the efficient choice. \poly{2} is the only learnable closure behind \phys{} ($+1.7$\,cm, not significant), and its $D_{\Mtwo}$ ranges over 2.18--3.31\,TW across folds.

\begin{table*}[t]
\centering\footnotesize
\caption{Five-fold random (hash) CV, fitted on $M_2,S_2,K_1,O_1$ (RSS over constituents, cm; mean\,$\pm$\,s.d.\ over folds). ``Pooled'' is the RMS over all 226 gauges. $\Delta$ vs.\ \phys{} is the pooled paired difference with its 95\% bootstrap CI (negative = better). $P(\text{model}\ge\text{GP})<2.5\times10^{-4}$ for every closure.}
\label{tab:cv}
\begin{tabular}{@{}lrcccccccc@{}}
\toprule
Closure & \#p & In-sample & Held-out & Gap & Pooled & $\Delta$ vs.\ \phys{} [95\% CI] & Wins & $D_{M_2}$ [TW] & Shelf [\%]\\
\midrule
\phys{}                              & 4       & $37.2\pm1.4$ & $37.7\pm5.5$ & $+0.5$ & 37.7 & --                           & --  & 2.55 & 21\\
\poly{2}                             & 11      & $35.5\pm1.1$ & $39.5\pm5.5$ & $+4.0$ & 39.4 & $+1.7\ [-1.4,+4.8]$          & 1/5 & 2.90 & 15\\
\nn{8}                               & 114     & $32.7\pm2.3$ & $34.5\pm3.3$ & $+1.9$ & 34.4 & $\mathbf{-3.3}\ [-5.7,-1.0]$ & 4/5 & 2.87 & 18\\
\nn{32}                              & 1\,218  & $30.6\pm1.8$ & $35.4\pm3.9$ & $+4.8$ & 35.3 & $\mathbf{-2.4}\ [-4.1,-0.7]$ & 4/5 & 3.02 & 20\\
\nn{128}                             & 17\,154 & $29.2\pm2.1$ & $35.1\pm5.7$ & $+6.0$ & 35.2 & $-2.5\ [-5.2,+1.0]$          & 4/5 & 3.00 & 20\\
\nn{32}[$H$, slope]                  & 1\,186  & $32.9\pm2.1$ & $36.8\pm8.4$ & $+3.8$ & 37.0 & $-0.7\ [-3.3,+2.0]$          & 2/5 & 2.83 & 21\\
\nn{32}[$H$, slope, $|\sin\varphi|$] & 1\,218  & $31.7\pm1.6$ & $\mathbf{34.0}\pm4.4$ & $+2.3$ & \textbf{33.9} & $\mathbf{-3.8}\ [-6.3,-1.6]$ & 5/5 & 2.99 & 22\\
\midrule
GP (data only)                       & --      & --           & $47.7\pm7.5$ & --     & 47.6 & --                           & --  & --   & --\\
\bottomrule
\end{tabular}
\end{table*}

\subsection{Spatial-Block Validation Separates Transferable Skill (E2, Axis A1)}
\label{sec:res:block}

\begin{figure*}[t]
\centering
\begin{minipage}[b]{0.43\textwidth}\centering
\figorbox{\linewidth}{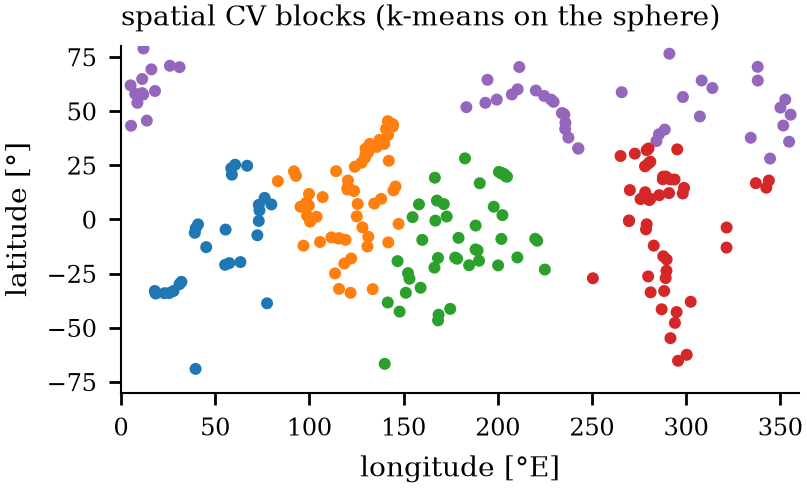}\\[-1mm]
{\footnotesize (a)}
\end{minipage}\hfill
\begin{minipage}[b]{0.55\textwidth}\centering
\figorbox{\linewidth}{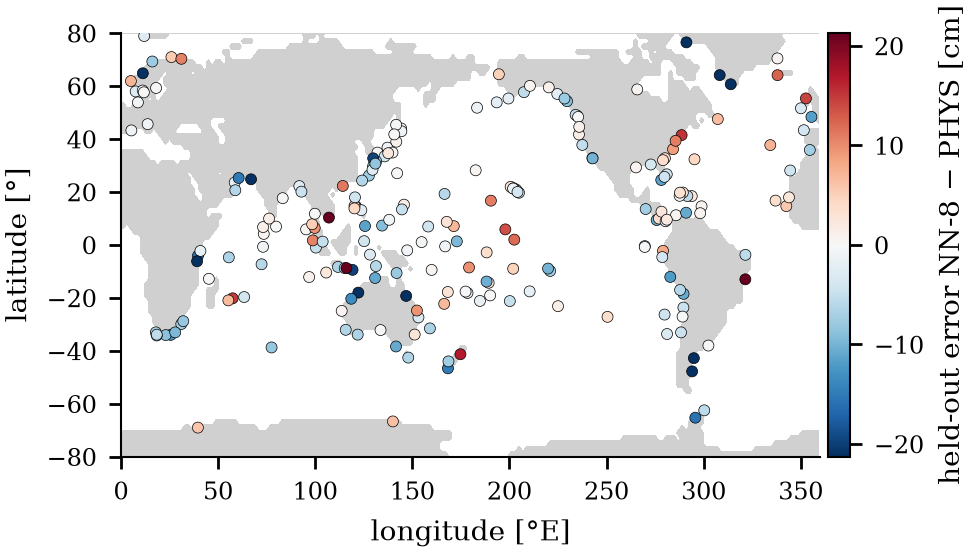}\\[-1mm]
{\footnotesize (b)}
\end{minipage}
\caption{(a)~Spatial CV blocks from spherical $k$-means: B0 Indian Ocean (blue), B1 western Pacific/East Asia/Australia (orange), B2 central Pacific (green), B3 Americas (red), B4 northern high latitudes (purple). (b)~Per-gauge held-out error difference \nn{8}$-$\phys{} under hash CV (blue: \nn{8} better). Gains cluster in the Southern Hemisphere (southern Africa, Patagonia, western Australia, Antarctica); losses cluster on the U.S.\ East Coast, in the South China Sea and in the north-east Atlantic.}
\label{fig:blocksgain}
\end{figure*}

The spherical $k$-means blocks (Fig.~\ref{fig:blocksgain}a) correspond to recognizable ocean regions:
\begin{itemize}
\item B0: Indian Ocean and southern Africa, 28 gauges;
\item B1: western Pacific, East Asia and Australia, 58;
\item B2: central and South Pacific islands, 46;
\item B3: the Americas and the western Atlantic, 46;
\item B4: northern high latitudes (Europe, North Atlantic and Alaska), 48.
\end{itemize}
Spatial blocking reveals how much of the random-CV gain is transferable (Table~\ref{tab:block}). Pooled over the \emph{same} 226 gauges, \nn{8} is 4.2\,cm [$+1.1,+7.8$] behind \phys{}, and \nn{32} 5.6\,cm [$+2.1,+9.5$] behind.

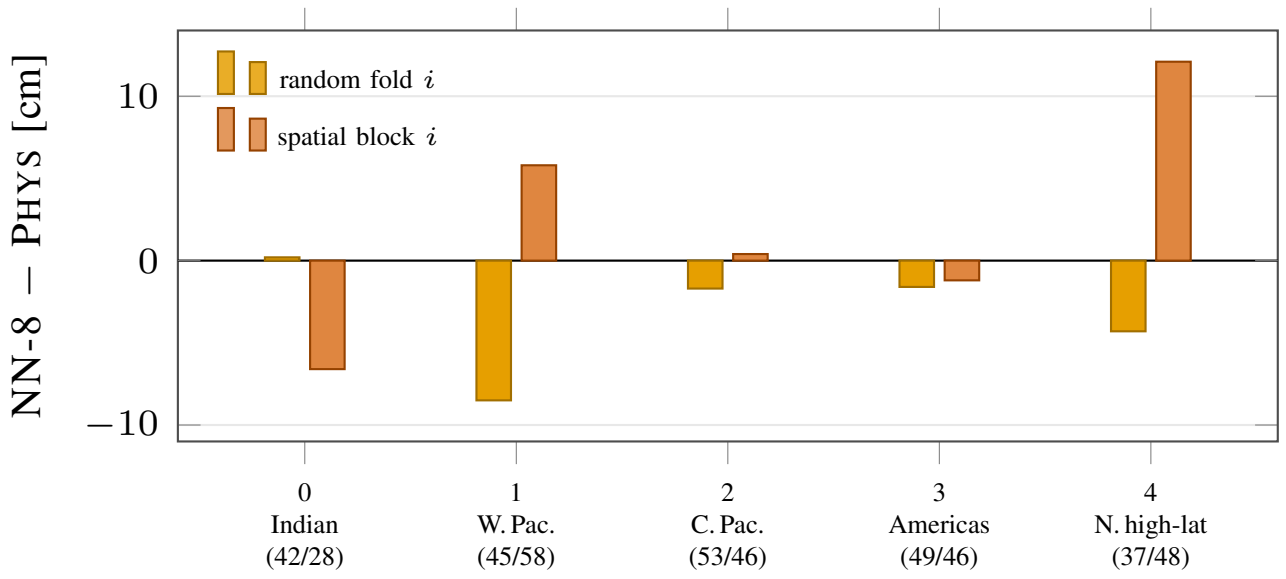
\begin{figure}[t]
\centering
\begin{tikzpicture}
\begin{axis}[hct, width=\columnwidth, height=4.3cm, ybar, bar width=6.5pt,
  ymin=-11, ymax=14, xtick={0,1,2,3,4},
  xticklabels={{0\\Indian\\(42/28)},{1\\W.\,Pac.\\(45/58)},{2\\C.\,Pac.\\(53/46)},{3\\Americas\\(49/46)},{4\\N.\,high-lat\\(37/48)}},
  xticklabel style={align=center, font=\tiny},
  ylabel={\nn{8} $-$ \phys{} [cm]},
  enlarge x limits=0.15, ymajorgrids,
  legend style={at={(0.02,0.98)}, anchor=north west, fill=none},
  extra y ticks={0}, extra y tick labels={}, extra y tick style={grid=major, grid style={black, thin}}]
\addplot[fill=cnn, draw=cnn!70!black] coordinates {(0,0.2)(1,-8.5)(2,-1.7)(3,-1.6)(4,-4.3)};
\addplot[fill=cred!75, draw=cred!70!black] coordinates {(0,-6.6)(1,5.8)(2,0.4)(3,-1.2)(4,12.1)};
\legend{random fold $i$, spatial block $i$}
\end{axis}
\end{tikzpicture}
\caption{Per-fold held-out difference between \nn{8} and \phys{} (negative: \nn{8} better). Region names refer to the spatial blocks; numbers in brackets are test gauges (hash/block). Under random folds \nn{8} leads in 4/5; under spatial blocks the comparison resolves by region, with the largest differences in the western-Pacific and northern high-latitude blocks.}
\label{fig:folddiff}
\end{figure}

Fig.~\ref{fig:folddiff} resolves this by region. \nn{8} leads in the Indian Ocean block by 6.6\,cm and trails in the western Pacific block by 5.8\,cm and in the northern high-latitude block by 12.1\,cm, the two regions where its learned drag is largest (Sec.~\ref{sec:res:maps}). When B4 is held out, almost no training gauges lie poleward of $\sim$50$^\circ$N, so this block directly tests whether a closure's northern drag is anchored in transferable physics.

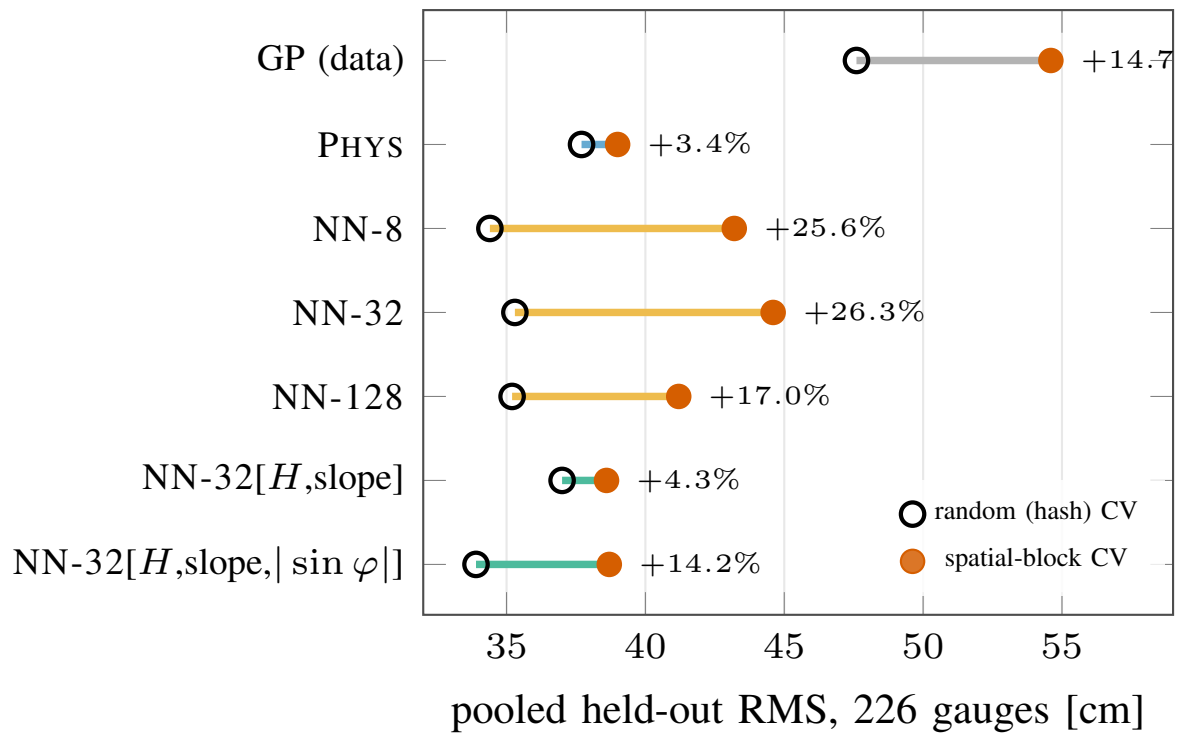
\begin{figure}[t]
\centering
\begin{tikzpicture}
\begin{axis}[hct, scale only axis, width=0.56\columnwidth, height=4.0cm,
  xmin=32, xmax=59, ymin=0.4, ymax=7.6,
  ytick={1,2,3,4,5,6,7},
  yticklabels={GP (data), \phys{}, \nn{8}, \nn{32}, \nn{128}, {\nn{32}[$H$,slope]}, {\nn{32}[$H$,slope,$|\sin\varphi|$]}},
  yticklabel style={font=\scriptsize},
  xtick={35,40,45,50,55},
  xlabel={pooled held-out RMS, 226 gauges [cm]},
  xmajorgrids, y dir=reverse,
  legend style={at={(0.99,0.03)}, anchor=south east}]
\draw[gray!60, line width=1.4pt] (axis cs:47.6,1)--(axis cs:54.6,1);
\draw[cphys!60, line width=1.4pt] (axis cs:37.7,2)--(axis cs:39.0,2);
\draw[cnn!70, line width=1.4pt] (axis cs:34.4,3)--(axis cs:43.2,3);
\draw[cnn!70, line width=1.4pt] (axis cs:35.3,4)--(axis cs:44.6,4);
\draw[cnn!70, line width=1.4pt] (axis cs:35.2,5)--(axis cs:41.2,5);
\draw[cgreen!70, line width=1.4pt] (axis cs:37.0,6)--(axis cs:38.6,6);
\draw[cgreen!70, line width=1.4pt] (axis cs:33.9,7)--(axis cs:38.7,7);
\addplot[only marks, mark=o, mark size=2.2pt, thick, black] coordinates {(47.6,1)(37.7,2)(34.4,3)(35.3,4)(35.2,5)(37.0,6)(33.9,7)};
\addplot[only marks, mark=*, mark size=2.2pt, cred] coordinates {(54.6,1)(39.0,2)(43.2,3)(44.6,4)(41.2,5)(38.6,6)(38.7,7)};
\legend{random (hash) CV, spatial-block CV}
\node[font=\tiny, anchor=west] at (axis cs:55.0,1) {$+14.7\%$};
\node[font=\tiny, anchor=west] at (axis cs:39.4,2) {$+3.4\%$};
\node[font=\tiny, anchor=west] at (axis cs:43.6,3) {$+25.6\%$};
\node[font=\tiny, anchor=west] at (axis cs:45.0,4) {$+26.3\%$};
\node[font=\tiny, anchor=west] at (axis cs:41.6,5) {$+17.0\%$};
\node[font=\tiny, anchor=west] at (axis cs:39.0,6) {$+4.3\%$};
\node[font=\tiny, anchor=west] at (axis cs:39.1,7) {$+14.2\%$};
\end{axis}
\end{tikzpicture}
\caption{Extrapolation penalty: pooled error on the same 226 gauges under random and spatial-block CV. The physical closure and the latitude-free MLP have the smallest penalties. The two MLPs without a signed latitude input (green) reach block-CV skill level with \phys{}; the hemispherically symmetric MLP does so from the best random-CV score, which is why its relative penalty ($+14.2\%$) is larger.}
\label{fig:dumbbell}
\end{figure}

Fig.~\ref{fig:dumbbell} summarizes the result as an \emph{extrapolation penalty}: the relative increase in pooled error from hash to block CV. \phys{} changes by only $+3.4\%$, against $+14.7\%$ for the GP and $+17$ to $+26\%$ for the MLPs with a signed latitude input. The value of the physics prior therefore \emph{grows} under extrapolation: \phys{} beats the GP by 29\% under block CV versus 21\% under hash CV.

The feature ablation isolates what fails to transfer. The latitude-free \nn{32}[$H$, slope] matches \phys{} under blocking ($-0.5$\,cm [$-2.5,+1.5$], penalty $+4.3\%$), but it gives up most of the random-CV gain. The hemispherically symmetric \nn{32}[$H$, slope, $|\sin\varphi|$], which has the same width and the same 1\,218 parameters as \nn{32}, keeps the best random-CV score (33.9\,cm) and is also statistically level with \phys{} under blocking (38.7\,cm, $-0.3$\,cm [$-3.6,+2.6$], 3/5 blocks won), with the lowest fold-mean held-out error of all closures ($34.6$\,cm). Exchanging $\sin\varphi$ for $|\sin\varphi|$ alone lowers the pooled block-CV error by 5.9\,cm. The symmetric network still pays a hash-to-block penalty of $+14.2\%$, because its random-CV score is the lowest of all; what the exchange removes is the deficit relative to \phys{}. Latitude information as such is therefore not what fails to transfer; the hemispheric sign is, because it lets a network store a north--south switch in the drag. Without a coordinate input that separates the hemispheres, the network is pushed toward encoding \emph{why} a region dissipates rather than \emph{where} it is. 
\begin{table}[t]
\centering\footnotesize
\caption{Spatial-block CV (5 spherical $k$-means blocks; same 226 gauges as Table~\ref{tab:cv}). In-sample and held-out: fold mean\,$\pm$\,s.d.\ [cm]; $\Delta$: pooled paired difference to \phys{} with 95\% bootstrap CI (negative = better); ``s'' denotes slope. GP pooled: 54.6\,cm.}
\label{tab:block}
\setlength{\tabcolsep}{3pt}
\begin{adjustbox}{max width=\columnwidth}
\begin{tabular}{@{}lccccc@{}}
\toprule
Closure & In-sample & Held-out & Pooled & $\Delta$ [95\% CI] & Wins\\
\midrule
\phys{}                        & $37.0\pm2.4$ & $36.5\pm11.2$ & 39.0 & --                                          & --\\
\nn{8}                         & $32.4\pm2.9$ & $38.5\pm16.7$ & 43.2 & $\mathbf{+4.2}$ {\scriptsize[$+1.1,+7.8$]}  & 2/5\\
\nn{32}                        & $30.7\pm1.7$ & $39.5\pm17.9$ & 44.6 & $\mathbf{+5.6}$ {\scriptsize[$+2.1,+9.5$]}  & 2/5\\
\nn{128}                       & $29.6\pm1.7$ & $36.5\pm16.2$ & 41.2 & $+2.1$ {\scriptsize[$-1.4,+6.0$]}           & 2/5\\
\nn{32}[$H$,s]                 & $33.2\pm2.2$ & $36.1\pm11.1$ & \textbf{38.6} & $-0.5$ {\scriptsize[$-2.5,+1.5$]}  & 2/5\\
\nn{32}[$H$,s,$|\sin\varphi|$] & $31.4\pm2.1$ & $\mathbf{34.6}\pm14.6$ & 38.7 & $-0.3$ {\scriptsize[$-3.6,+2.6$]} & 3/5\\
\bottomrule
\end{tabular}
\end{adjustbox}
\end{table}

\subsection{Atlas Benchmark: Representativeness Floor and Gauge Strata (E8)}
\label{sec:res:bench}

\begin{figure}[t]
\centering
\begin{tikzpicture}
\begin{axis}[hct, width=\columnwidth, height=3.7cm, ybar, bar width=4.5pt,
  ymin=0, ymax=46, symbolic x coords={P,N8,N32,NA,NH,GP}, xtick=data,
  xticklabels={\phys{}, \nn{8}, \nn{32}, {\nn{32}\\ $[|\sin\varphi|]$}, {\nn{32}\\ $[H,\mathrm{slope}]$}, GP},
  xticklabel style={align=center, font=\tiny},
  ylabel={RMS [cm]}, enlarge x limits=0.1, ymajorgrids,
  legend columns=2, legend style={at={(0.02,0.98)}, anchor=north west},
  title={(a) open-ocean gauges ($n=128$)}, title style={font=\scriptsize},
  extra y ticks={8.7}, extra y tick labels={}, extra y tick style={grid=major, grid style={cgreen, densely dashed, thick}}]
\addplot[fill=cphys!70, draw=none] coordinates {(P,21.6)(N8,20.6)(N32,18.9)(NA,19.4)(NH,20.1)(GP,26.7)};
\addplot[fill=cred!75, draw=none] coordinates {(P,22.4)(N8,21.6)(N32,23.2)(NA,21.6)(NH,22.6)(GP,39.6)};
\legend{hash CV, block CV}
\node[font=\tiny, text=cgreen!60!black, anchor=south east] at (rel axis cs:0.99,0.19) {$2^\circ$ floor 8.7};
\end{axis}
\end{tikzpicture}

\vspace{1mm}
\begin{tikzpicture}
\begin{axis}[hct, width=\columnwidth, height=3.7cm, ybar, bar width=4.5pt,
  ymin=0, ymax=80, symbolic x coords={P,N8,N32,NA,NH,GP}, xtick=data,
  xticklabels={\phys{}, \nn{8}, \nn{32}, {\nn{32}\\ $[|\sin\varphi|]$}, {\nn{32}\\ $[H,\mathrm{slope}]$}, GP},
  xticklabel style={align=center, font=\tiny},
  ylabel={RMS [cm]}, enlarge x limits=0.1, ymajorgrids,
  title={(b) coastal gauges ($n=93$)}, title style={font=\scriptsize},
  extra y ticks={35.4}, extra y tick labels={}, extra y tick style={grid=major, grid style={cgreen, densely dashed, thick}}]
\addplot[fill=cphys!70, draw=none] coordinates {(P,55.3)(N8,49.9)(N32,52.6)(NA,49.7)(NH,55.0)(GP,70.2)};
\addplot[fill=cred!75, draw=none] coordinates {(P,57.2)(N8,65.2)(N32,66.5)(NA,57.2)(NH,56.2)(GP,73.1)};
\node[font=\tiny, text=cgreen!60!black, anchor=south west] at (rel axis cs:0.01,0.445) {$2^\circ$ floor 35.4};
\end{axis}
\end{tikzpicture}
\caption{Gauge strata on the 221 benchmark gauges; dashed green lines mark the GOT4.10c representativeness floor at $2^\circ$. In the open ocean every closure stays far below the GP under both designs; the learned closures are below \phys{} under hash CV, and \nn{8} and the hemispherically symmetric network remain slightly below it under block CV. The extrapolation penalty of the signed-latitude networks (\nn{8}, \nn{32}) is concentrated at coastal gauges; under block CV the hemispherically symmetric and latitude-free networks stay at the coastal level of \phys{}.}
\label{fig:strata}
\end{figure}
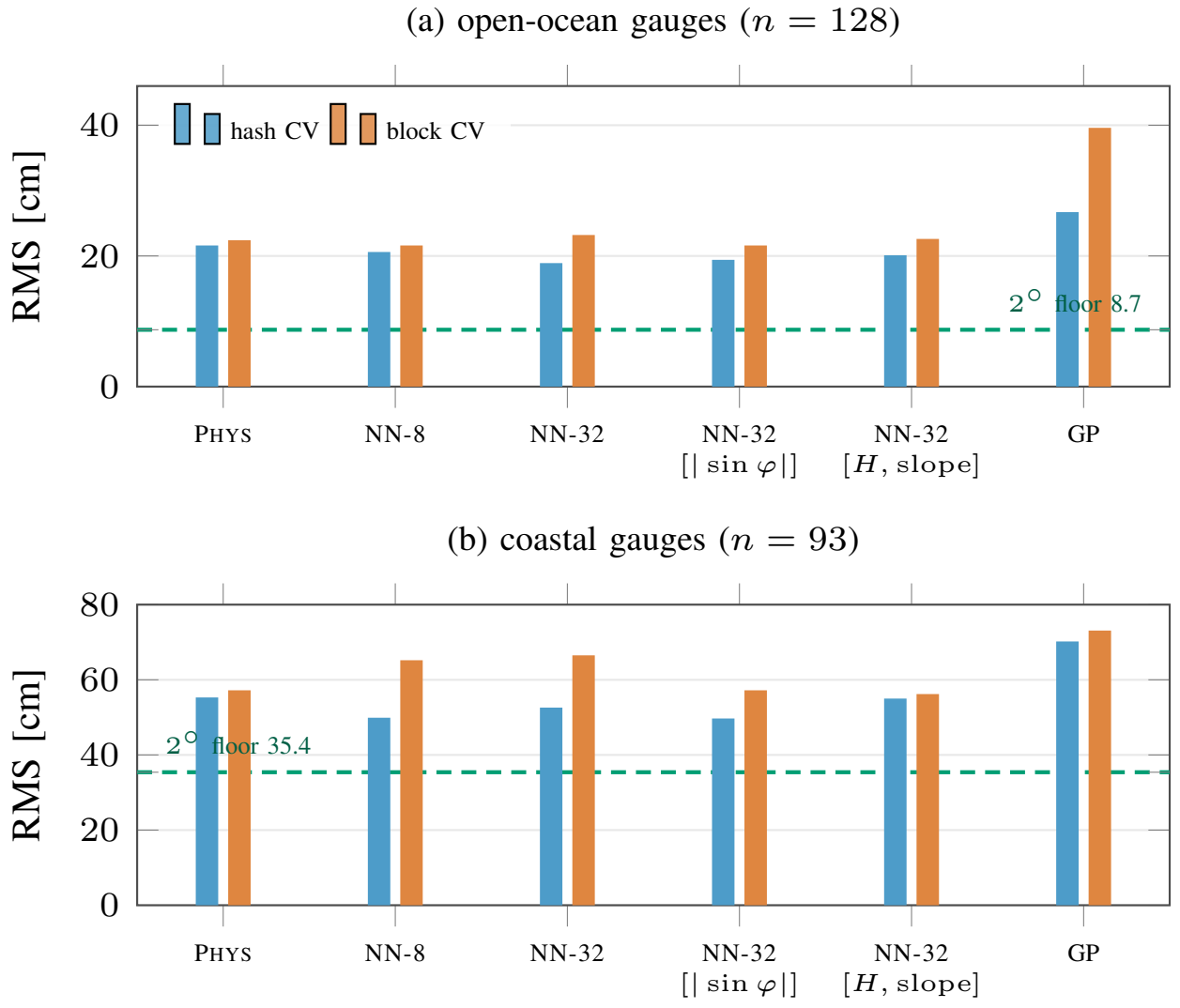

Table~\ref{tab:bench} and Fig.~\ref{fig:strata} place the gauge scores on an absolute scale. Evaluated on its native grid, GOT4.10c reproduces the gauge constants to 9.9\,cm (5.4\,cm open, 14.7\,cm coastal). The closures studied here are therefore not meant to compete with data-assimilative atlases in absolute accuracy at these resolutions; the comparison of interest is between closures under identical conditions. Sampled at the $2^\circ$ cell that the gauge operator uses, the same atlas scores 22.8\,cm (8.7\,cm open, 35.4\,cm coastal). This representativeness floor is common to all models. In RMS terms it equals 60\% of the pooled error of \phys{} (38.0\,cm on these 221 gauges) and 67\% of that of the best closure (34.1\,cm); because independent errors add in quadrature, the corresponding shares of error variance are 36\% and 45\%. Measured as excess over the floor, i.e., as a difference of pooled RMS values, \phys{} is $+15.2$\,cm [$+11.2,+19.6$] above it and the GP $+25.2$\,cm [$+18.7,+31.7$]. On this measure the physics-constrained closures remove 33--55\% of the GP's excess under hash CV (31--50\% under block CV; 23--40\% under hash CV if the excess is taken in quadrature), more than the 17--29\% reduction in total error suggests.

Stratification yields three further results.

\emph{(i) The physics prior matters most away from the coasts.} Under block CV the GP's open-ocean error rises from 26.7 to 39.6\,cm, whereas that of \phys{} rises only from 21.6 to 22.4\,cm, so \phys{} is 43\% below the GP in the open ocean under extrapolation.

\emph{(ii) The neural closures' extrapolation penalty is mainly coastal.} For \nn{8}, the coastal error rises from 49.9 to 65.2\,cm ($+31\%$) between hash and block CV, while the open-ocean error rises only from 20.6 to 21.6\,cm ($+5\%$), which leaves \nn{8} about 1\,cm below \phys{} in the open ocean under both designs (pooled strata scores, reported without an interval). For \nn{32} and \nn{128} the penalty appears in both strata, but coastal gauges dominate in absolute terms ($+13.9$ and $+9.2$\,cm vs.\ $+4.3$ and $+2.4$\,cm). The hemispherically symmetric network equals \phys{} at the coast under blocking (57.2\,cm for both) and is slightly below it in the open ocean (21.6 vs.\ 22.4\,cm).

\emph{(iii) The open-ocean gains are also visible in the field, not only at the gauges.} Over deep cells, every learned closure is closer to the atlas than \phys{} (12.5--14.2 vs.\ 15.4\,cm), and the distance falls monotonically with width (\nn{8} 14.0, \nn{32} 13.0, \nn{128} 12.5\,cm). Over shallow cells all closures differ from the atlas by 45--50\,cm, a level set by sub-grid shelf seas; the latitude-free network and \poly{2} are furthest from it (49.7 and 49.0\,cm).

Together these results split the learned closures' random-CV lead into two parts: an improvement of the resolved, open-ocean tide that transfers across regions and is supported by the atlas, and a coastal correction that holds only within the region it was trained on. The next section examines the physical origin of the second part.

\begin{table*}[t]
\centering\footnotesize
\caption{Atlas benchmark on the 221 gauges with valid GOT4.10c values (128 open, snapped-cell depth $\ge1\,000$\,m; 93 coastal). RSS over $M_2,S_2,K_1,O_1$ in cm. ``Excess'' is the pooled paired difference to the $2^\circ$ representativeness floor with its 95\% bootstrap CI. ``Field'' is the RSS difference between the fitted solution and the atlas over deep and shallow $2^\circ$ wet cells (hash-CV models).}
\label{tab:bench}
\setlength{\tabcolsep}{3.2pt}
\begin{tabular}{@{}lccccccccc@{}}
\toprule
 & \multicolumn{3}{c}{Hash CV} & \multicolumn{3}{c}{Block CV} & \multicolumn{2}{c}{Excess over floor, all} & Field vs.\ atlas\\
\cmidrule(lr){2-4}\cmidrule(lr){5-7}\cmidrule(lr){8-9}\cmidrule(l){10-10}
Model & all & open & coastal & all & open & coastal & hash & block & deep / shallow\\
\midrule
\phys{}                              & 38.0 & 21.6 & 55.3 & 39.3 & 22.4 & 57.2 & $+15.2\ [+11.2,+19.6]$ & $+16.5\ [+12.4,+21.0]$ & 15.4 / 45.7\\
\poly{2}                             & 39.8 & 21.2 & 58.8 & --   & --   & --   & $+17.0\ [+12.6,+21.5]$ & --                     & 14.2 / 49.0\\
\nn{8}                               & 34.7 & 20.6 & 49.9 & 43.5 & 21.6 & 65.2 & $+11.9\ [+8.5,+16.1]$  & $+20.8\ [+15.7,+26.4]$ & 14.0 / 45.2\\
\nn{32}                              & 35.6 & 18.9 & 52.6 & 44.7 & 23.2 & 66.5 & $+12.8\ [+8.4,+17.7]$  & $+22.0\ [+16.6,+27.9]$ & 13.0 / \textbf{45.1}\\
\nn{128}                             & 35.5 & \textbf{18.5} & 52.7 & 41.4 & \textbf{20.9} & 61.9 & $+12.7\ [+7.9,+18.3]$  & $+18.6\ [+13.7,+24.4]$ & \textbf{12.5} / 45.9\\
\nn{32}[$H$,slope]                   & 37.3 & 20.1 & 55.0 & \textbf{38.8} & 22.6 & \textbf{56.2} & $+14.5\ [+10.0,+19.6]$ & $\mathbf{+16.1}\ [+11.9,+20.7]$ & 13.7 / 49.7\\
\nn{32}[$H$,slope,$|\sin\varphi|$]   & \textbf{34.1} & 19.4 & \textbf{49.7} & 39.0 & 21.6 & 57.2 & $\mathbf{+11.4}\ [+7.2,+16.1]$ & $+16.2\ [+11.7,+20.9]$ & 13.7 / 46.8\\
\midrule
GP (data only)                       & 48.0 & 26.7 & 70.2 & 54.7 & 39.6 & 73.1 & $+25.2\ [+18.7,+31.7]$ & $+31.9\ [+26.9,+37.6]$ & --\\
Equilibrium tide                     & 56.5 & 42.1 & 74.4 & 56.5 & 42.1 & 74.4 & $+33.7\ [+28.8,+38.9]$ & $+33.7\ [+28.8,+38.9]$ & 35.3 / 57.2\\
\midrule
GOT4.10c, native grid                & 9.9  & 5.4  & 14.7 & 9.9  & 5.4  & 14.7 & -- & -- & --\\
GOT4.10c at $2^\circ$ cell (floor)   & 22.8 & 8.7  & 35.4 & 22.8 & 8.7  & 35.4 & 0  & 0  & --\\
\bottomrule
\end{tabular}
\end{table*}

\subsection{What the Network Learns: A Diagnostic of Dissipation at Northern Shelf-Sea Latitudes}
\label{sec:res:maps}

\begin{figure*}[t]
\centering
\figorbox{0.92\textwidth}{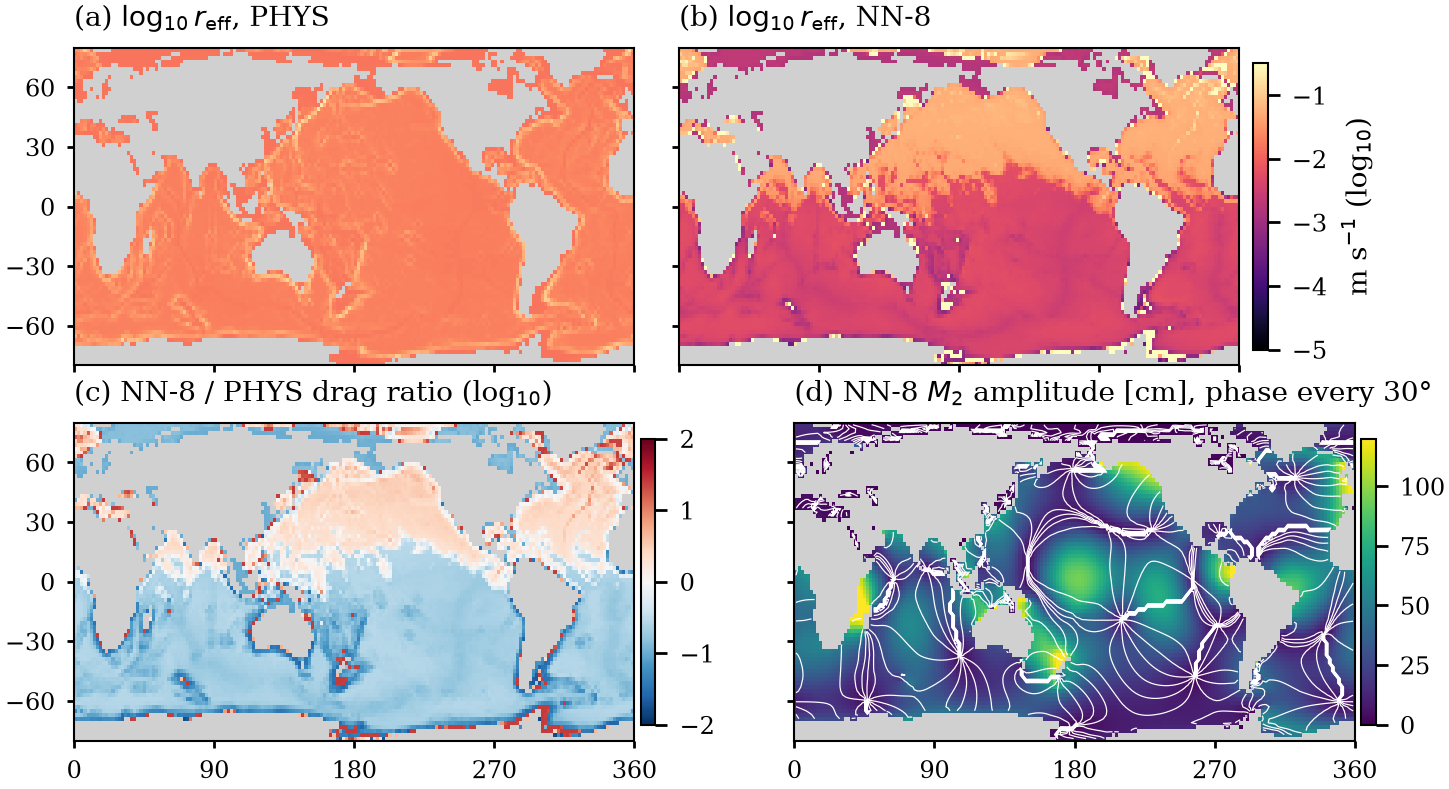}
\caption{Learned dissipation fields (fold 0, hash CV). (a)~$\log_{10}r_{\mathrm{eff}}$ of \phys{}; (b)~$\log_{10}r_{\mathrm{eff}}$ of \nn{8}; (c)~$\log_{10}$ ratio \nn{8}/\phys{}, with the latitude switch near 15$^\circ$N; (d)~\nn{8} $M_2$ amplitude with co-phase lines every $30^\circ$. The amphidromic system is preserved while the drag field is organized by latitude rather than by bathymetry, which is why the framework scores closures on dissipation diagnostics in addition to gauge skill.}
\label{fig:maps}
\end{figure*}

The learned fields help explain the difference between the two validation designs and, read through the framework's diagnostics, suggest a specific physical deficit. The \phys{} drag (Fig.~\ref{fig:maps}a) is nearly uniform, with enhancement over rough topography such as mid-ocean ridges and island arcs. The \nn{8} drag (Fig.~\ref{fig:maps}b,c) is organized by latitude: about one order of magnitude above \phys{} north of $\sim$15$^\circ$N (North Pacific and North Atlantic) and about one order below it across the Southern Hemisphere, with its shelf cells at the upper bound of the constraint layer.

\begin{figure}[t]
\centering
\figorbox{\columnwidth}{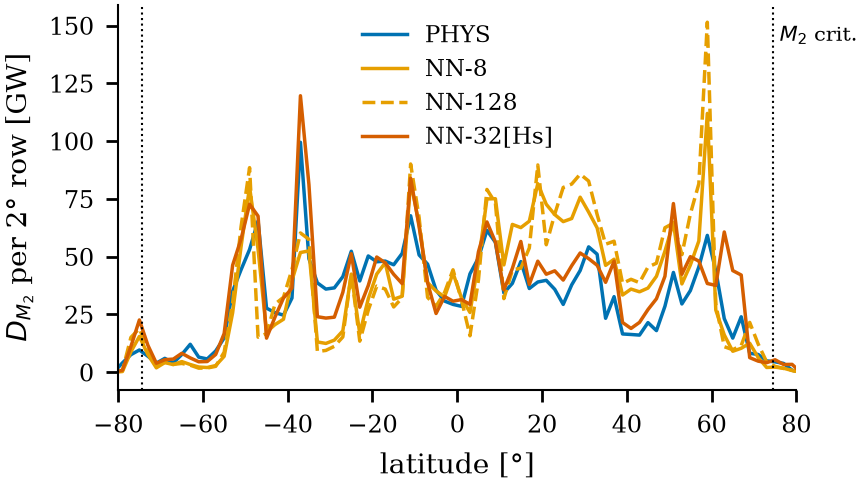}
\caption{Zonal $M_2$ dissipation per $2^\circ$ latitude row (hash CV, fold mean) for \phys{}, \nn{8}, \nn{128} (dashed) and the latitude-free \nn{32}[$H$,slope]. Dotted lines: $M_2$ critical latitude. The latitude-aware MLPs add dissipation in 15--35$^\circ$N and near 58$^\circ$N, the latitude bands of the major northern shelf seas.}
\label{fig:zonal}
\end{figure}

The zonal dissipation budget (Fig.~\ref{fig:zonal}) shows where the additional energy goes. Relative to \phys{}, the MLPs place $\Mtwo$ dissipation in 15--35$^\circ$N and in a sharp peak near 58$^\circ$N (up to 150\,GW per $2^\circ$ row for \nn{128}). These bands are consistent with the principal northern shelf-sea sinks of $\Mtwo$ energy identified from altimetry~\cite{egbert2001}: the Yellow and East China Seas at 25--40$^\circ$N, and the European shelf, Hudson Bay, the Sea of Okhotsk and the Bering Sea at 50--65$^\circ$N. These seas are largely sub-grid at $2^\circ$, which is also why every closure's shelf share (15--22\%) stays below the observed two-thirds (Sec.~\ref{sec:res:energy}). A plausible reading is that the network compensates for this missing shelf dissipation and, because the signed latitude is the feature that best separates these seas at this resolution, stores the correction as a hemispheric switch. The added dissipation sits at these latitudes rather than on the resolved shelf cells: the shelf share of \nn{8} (18\%) is no higher than that of \phys{} (21\%). Three independent diagnostics are consistent with this reading. Southern Hemisphere gauges gain from lower drag (Fig.~\ref{fig:blocksgain}b), and the northern block, where the correction is concentrated, is the one that most tests its transfer (Fig.~\ref{fig:folddiff}). The block-CV penalty falls on coastal rather than open-ocean gauges (Sec.~\ref{sec:res:bench}), as expected for a correction that stands in for sub-grid shelf response. Finally, removing the hemispheric sign from the input ($\sin\varphi\to|\sin\varphi|$) brings the block-CV error back to the level of \phys{} (Sec.~\ref{sec:res:block}). Section~\ref{sec:res:res} tests a prediction of this reading by refining the grid.

The framework turns this diagnostic into action in three ways: the hemispherically symmetric closure and the latitude-free closure each match \phys{} under blocking (Sec.~\ref{sec:res:block}), and the energy penalty moves the network's dissipation onto the shelf (Sec.~\ref{sec:res:energy}). A neural closure is therefore most valuable here as a \emph{diagnostic}: its gain map and zonal budget indicate where the physical closure may benefit from additional shelf physics.

\subsection{Regularization as a Control Dial (E3)}
\label{sec:res:reg}

Weak weight decay ($\lambda=10^{-4}$) gives the best width-8 model (34.4\,cm) and narrows the gap at larger widths (Table~\ref{tab:reg}). Strong decay ($\lambda=10^{-3}$) collapses all widths onto practically the same solution: their held-out errors coincide to within 0.5\,cm in every fold, e.g., 44.4\,cm in fold~1 for all widths against 44.1\,cm for \phys{}, and overall skill returns to \phys{} level (37.6--37.8\,cm). Weight decay is therefore a dial that moves the learned closure smoothly between the regional correction of Sec.~\ref{sec:res:maps} and the physical closure. It also indicates that the $\approx3$\,cm random-CV advantage of the MLPs is carried by that spatially structured drag.

\begin{table}[t]
\centering\footnotesize
\caption{Weight-decay sweep under hash CV (fold means, cm). The $\lambda=10^{-6}$ rows are the E1 fits of Table~\ref{tab:cv}.}
\label{tab:reg}
\setlength{\tabcolsep}{4pt}
\begin{tabular}{@{}lccccc@{}}
\toprule
$\lambda$ & Closure & In-sample & Held-out & Gap & $D_{M_2}$ [TW]\\
\midrule
\multirow{3}{*}{$10^{-6}$} & \nn{8}   & 32.7 & 34.5 & $+1.9$ & 2.87\\
                           & \nn{32}  & 30.6 & 35.4 & $+4.8$ & 3.02\\
                           & \nn{128} & 29.2 & 35.1 & $+6.0$ & 3.00\\
\midrule
\multirow{3}{*}{$10^{-4}$} & \nn{8}   & 32.5 & \textbf{34.4} & $+1.9$ & 2.85\\
                           & \nn{32}  & 31.2 & 34.7 & $+3.4$ & 3.01\\
                           & \nn{128} & 30.2 & 35.1 & $+4.8$ & 2.95\\
\midrule
\multirow{3}{*}{$10^{-3}$} & \nn{8}   & 36.2 & 37.6 & $+1.4$ & 2.59\\
                           & \nn{32}  & 35.9 & 37.7 & $+1.8$ & 2.74\\
                           & \nn{128} & 35.9 & 37.8 & $+1.9$ & 2.73\\
\bottomrule
\end{tabular}
\end{table}

\subsection{Frequency Transfer (E4, Axis A2)}
\label{sec:res:transfer}

\begin{figure}[t]
\centering
\begin{tikzpicture}
\begin{axis}[hct, width=\columnwidth, height=4.4cm, ybar, bar width=5.2pt,
  ymin=0.6, ymax=1.8,
  symbolic x coords={M2,S2,K1,O1}, xtick=data,
  xticklabels={{$M_2$ (fitted)},{$S_2$},{$K_1$},{$O_1$}},
  ylabel={held-out RMS / GP},
  enlarge x limits=0.18, ymajorgrids,
  legend columns=2, legend style={at={(0.02,0.98)}, anchor=north west},
  extra y ticks={1}, extra y tick labels={}, extra y tick style={grid=major, grid style={black, densely dashed}}]
\addplot[fill=cphys, draw=none] coordinates {(M2,0.780)(S2,0.852)(K1,0.824)(O1,0.971)};
\addplot[fill=cnn, draw=none] coordinates {(M2,0.775)(S2,0.852)(K1,1.010)(O1,1.443)};
\addplot[fill=cred, draw=none] coordinates {(M2,0.735)(S2,0.858)(K1,1.127)(O1,1.686)};
\addplot[fill=cgreen, draw=none] coordinates {(M2,0.745)(S2,0.852)(K1,1.049)(O1,1.571)};
\legend{\phys{}, \nn{8}, {\nn{8}[$|\sin\varphi|$]}, {\nn{8}[$|f|/\omega$]}}
\node[font=\tiny, anchor=south east] at (rel axis cs:0.99,0.33) {GP = 1};
\end{axis}
\end{tikzpicture}
\caption{Frequency transfer, shown as held-out RMS relative to the data-only GP (dashed line; lower is better). All closures are fitted to $M_2$ only. The physical closure transfers to all four constituents; the learned closures transfer within the semidiurnal band, and their largest $M_2$ gains are the most frequency-specific.}
\label{fig:transfer}
\end{figure}
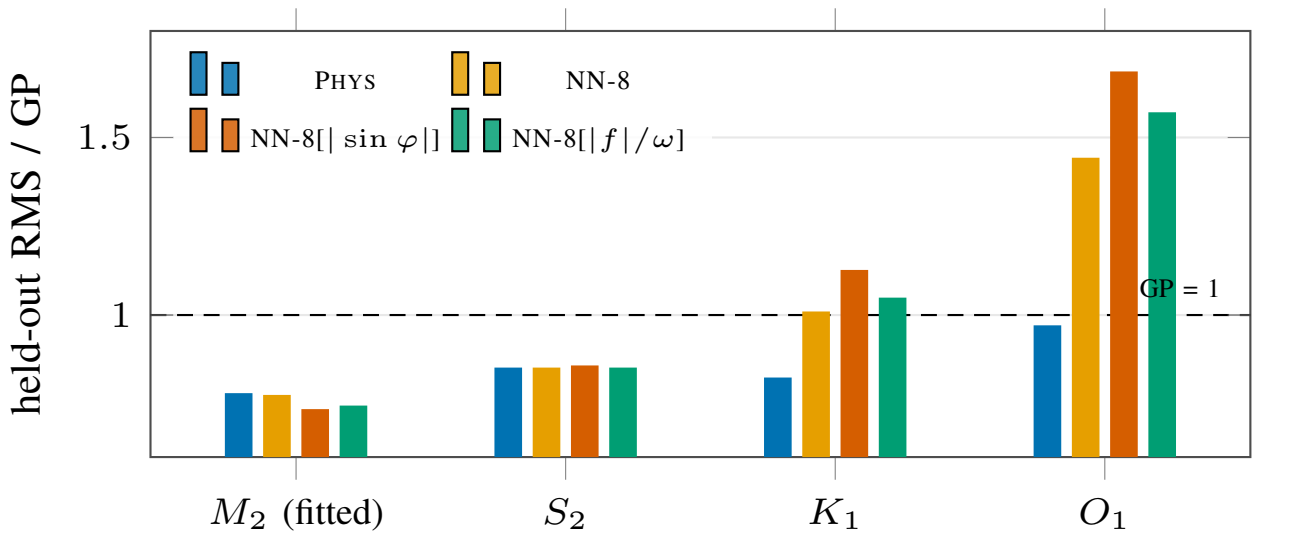

Frequency transfer is the cheapest test of whether a closure encodes \emph{physics} or a \emph{frequency-specific} correction. All closures are fitted to $M_2$ alone and scored on $S_2$, $K_1$ and $O_1$ without refitting (Table~\ref{tab:transfer}, Fig.~\ref{fig:transfer}).

\phys{} is frequency-portable across the four constituents tested. Fitted to $M_2$ alone, it reaches a four-constituent RSS of 37.7\,cm, identical to its four-constituent fit, and it is below the GP on both diurnal constituents (clearly on $K_1$, narrowly on $O_1$: 6.8 vs.\ 7.0\,cm). A compact physical law calibrated on a single constituent through the framework thus already reproduces the four major constituents at the accuracy of the joint fit. \nn{8} matches \phys{} on $M_2$ and $S_2$, so semidiurnal transfer is preserved, while diurnal skill changes by 23\% ($K_1$) and 49\% ($O_1$), with $O_1$ at 10.1\,cm against 7.0\,cm for the GP. The audit thus separates the semidiurnal component of the learned correction, which transfers, from a diurnal component that is specific to the $M_2$ fit. Adding a frequency-aware feature ($|f|/\omega$) or $|\sin\varphi|$ produces the best $M_2$ scores and shows the same diurnal pattern (up to $1.69\times$ the GP on $O_1$), which supports fitting all four constituents jointly, as in E1, as the training protocol for learned closures. In particular, the hemispherically symmetric feature that matches \phys{} under spatial blocking (Sec.~\ref{sec:res:block}) shows the weakest diurnal transfer when only $M_2$ is fitted, so spatial and frequency transfer are distinct axes that must be audited separately. The critical-latitude cut-off of \physfc{} has no measurable effect at this resolution, so the simpler \phys{} is sufficient.

\begin{table}[t]
\centering\footnotesize
\caption{Frequency transfer: fitted on $M_2$ only, scored on held-out gauges without refitting (hash CV, fold means, cm). RSS-4: root-sum-square over the four constituents.}
\label{tab:transfer}
\setlength{\tabcolsep}{3pt}
\begin{adjustbox}{max width=\columnwidth}
\begin{tabular}{@{}lcccccc@{}}
\toprule
Closure & $M_2$ (fit) & $S_2$ & $K_1$ & $O_1$ & RSS-4 & $D_{M_2}$\\
\midrule
\phys{}                           & 33.6 & 13.2 & \textbf{8.4} & \textbf{6.8} & \textbf{37.7} & 2.54\\
\physfc{}                         & 33.6 & 13.2 & 8.6 & 7.0 & 37.8 & 2.54\\
\nn{8}                            & 33.4 & 13.2 & 10.3 & 10.1 & 38.7 & 3.00\\
\nn{8}[$H$,slope,$|\sin\varphi|$] & \textbf{31.7} & 13.3 & 11.5 & 11.8 & 38.1 & 2.75\\
\nn{8}[$H$,slope,$|f|/\omega$]    & 32.1 & 13.2 & 10.7 & 11.0 & 37.9 & 2.85\\
\midrule
GP (data only)                    & 43.1 & 15.5 & 10.2 & 7.0 & 47.4 & --\\
\bottomrule
\end{tabular}
\end{adjustbox}
\end{table}

\subsection{Energy Admissibility (E5, E5b, Axis A3)}
\label{sec:res:energy}

Without an energy penalty, the learned closures dissipate 2.83--3.02\,TW of $M_2$ energy (18--26\% above the altimetric 2.4\,TW), while \phys{} stays within about 6\% (2.55\,TW). Part of the MLPs' gauge-fit gain is therefore associated with additional dissipation, located at the latitudes of the northern shelf seas identified in Sec.~\ref{sec:res:maps}. All closures place 15--22\% of dissipation on the shelf, below the observed $\approx2/3$, because most shelf seas are sub-grid at $2^\circ$. This remains so at $1^\circ$, where the shelf share is 19--25\% and the unconstrained fits dissipate 2.92--3.14\,TW (Sec.~\ref{sec:res:res}).

\begin{figure}[t]
\centering
\resizebox{\columnwidth}{!}{%
\begin{tikzpicture}
\begin{axis}[hct, name=ep1, width=5.2cm, height=4.4cm,
  xmin=-0.01, xmax=0.215, ymin=33.5, ymax=42.5,
  xlabel={$|D_{M_2}/2.4\,\mathrm{TW}-1|$}, ylabel={held-out RMS [cm]},
  grid=major, legend style={at={(0.98,0.98)}, anchor=north east}]
\addplot[cphys, thick, mark=*, mark size=1.8pt] coordinates {(0.0625,37.7)(0.054,37.8)(0.0375,37.8)(0.0042,38.5)(0.0625,41.3)};
\addplot[cnn, thick, mark=*, mark size=1.8pt] coordinates {(0.196,34.5)(0.158,35.0)(0.121,36.2)(0.050,40.3)(0.0083,39.3)};
\addplot[only marks, mark=star, mark size=3pt, cnnd, thick] coordinates {(0.0083,36.7)};
\legend{\phys{}, \nn{8}, {\nn{8}, budget only}}
\node[font=\tiny, anchor=south west] at (axis cs:0.196,34.5) {0};
\node[font=\tiny, anchor=south west] at (axis cs:0.158,35.0) {.03};
\node[font=\tiny, anchor=south west] at (axis cs:0.121,36.2) {.1};
\node[font=\tiny, anchor=south west] at (axis cs:0.050,40.3) {.3};
\node[font=\tiny, anchor=south west] at (axis cs:0.0083,39.3) {1};
\node[font=\tiny, anchor=south east, cphys] at (axis cs:0.0625,41.3) {1};
\node[font=\tiny, anchor=north west, cphys] at (axis cs:0.0625,37.7) {0};
\end{axis}
\begin{axis}[hct, at={(ep1.east)}, anchor=west, xshift=1.3cm, width=5.2cm, height=4.4cm,
  xmode=log, xmin=0.007, xmax=1.4, ymin=10, ymax=72,
  xlabel={$\mu$ (0 plotted at 0.01)}, ylabel={\% of $D_{M_2}$ in $H<500$\,m}, grid=major]
\addplot[cphys, thick, mark=*, mark size=1.8pt] coordinates {(0.01,21)(0.03,22)(0.1,25)(0.3,32)(1,48)};
\addplot[cnn, thick, mark=*, mark size=1.8pt] coordinates {(0.01,18)(0.03,22)(0.1,26)(0.3,38)(1,57)};
\addplot[cgreen, densely dashed] coordinates {(0.007,66.7)(1.4,66.7)};
\node[font=\tiny, cgreen!60!black, anchor=south west] at (axis cs:0.008,66.7) {observed $\approx2/3$};
\end{axis}
\end{tikzpicture}}
\caption{Energy-constrained fits. Left: skill against relative $D_{M_2}$ error (labels: $\mu$). At $\mu=1$, \nn{8} is both more accurate and closer to 2.4\,TW than \phys{}, while \phys{} is closest to the target at $\mu=0.3$; with the global-budget term alone (star) \nn{8} reaches the target at a fold-mean error about 1\,cm below unconstrained \phys{}. Right: shelf share of dissipation against $\mu$; the flexible closure moves furthest toward the observed value.}
\label{fig:pareto}
\end{figure}
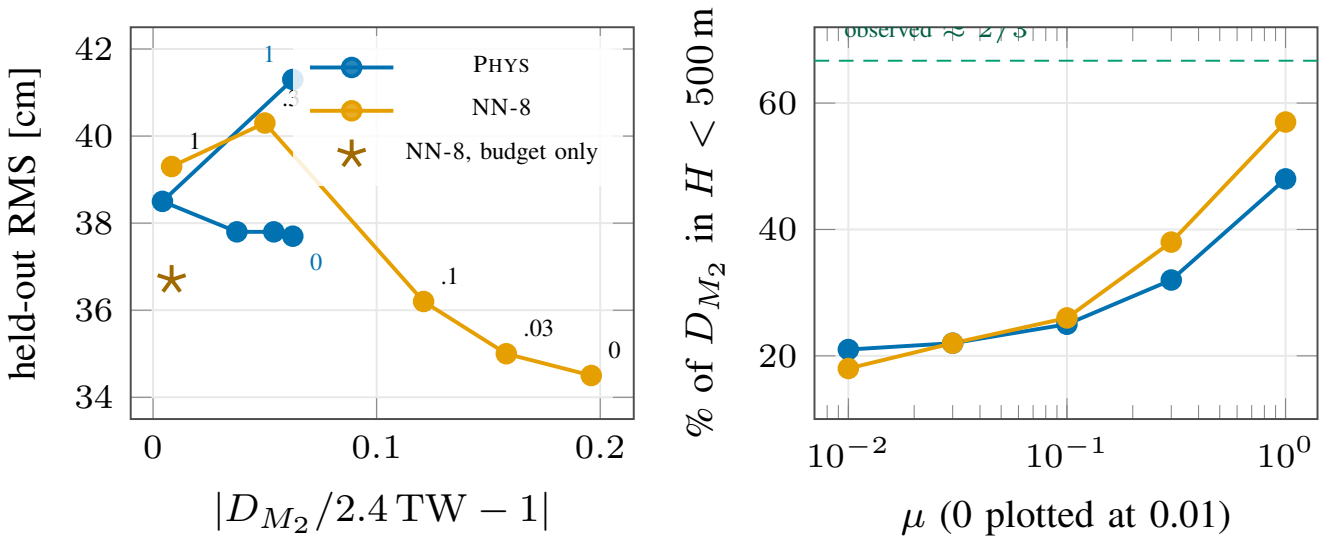

Because $D$ and $f_{\mathrm{sh}}$ are differentiable by-products of the solve, the framework lets admissibility be dialled in without extra solves (Table~\ref{tab:energy}, Fig.~\ref{fig:pareto}). A light penalty ($\mu=0.03$) keeps most of the \nn{8} gain (35.0\,cm) while lowering $D_{\Mtwo}$ to 2.78\,TW. At $\mu=1$, \nn{8} reaches 2.38\,TW, within 1\% of the target, and a 57\% shelf share at 39.3\,cm, whereas \phys{} reaches 2.25\,TW and 48\% at 41.3\,cm. \phys{} meets the budget target at a lighter penalty (2.39\,TW at 38.5\,cm for $\mu=0.3$), so the distinct advantage of the flexible closure is the shelf share, where it moves furthest toward the observed two-thirds. Capacity that random CV spends on a regional correction is thus redirected by the energy constraint toward the shelf, where the observations place the dissipation. This constructive use of capacity is visible only because the framework exposes the energy budget as a differentiable quantity. The \nn{8} response is not monotone in $\mu$: at $\mu=0.3$ its error (40.3\,cm) exceeds that at $\mu=1$ (39.3\,cm), consistent with the penalty reshaping the optimization landscape so that, within the fixed evaluation budget, intermediate coupling reaches a less favourable compromise.

The term ablation (Table~\ref{tab:ablation}) shows which of the two targets costs skill. The \emph{global budget alone} ($\mu_D=1$, $\mu_{\mathrm{sh}}=0$) is nearly free. It brings \nn{8} to 2.42\,TW, within 1\% of the target it is constrained toward, at 36.7\,cm, which is still about 1\,cm below unconstrained \phys{} in fold mean (37.7\,cm; \nn{8} wins 3/5 folds), a margin we regard as indicative, and it even improves the open-ocean error (20.6$\to$20.0\,cm). Its cost falls on coastal gauges (49.3$\to$53.7\,cm), consistent with the over-dissipation having served as a stand-in for sub-grid coastal response. For \phys{} the same term changes the error by only 0.1\,cm. The \emph{shelf-share term} is what costs skill: adding it raises the open-ocean error of both closures (\phys{} 21.5$\to$26.3\,cm, \nn{8} 20.0$\to$23.3\,cm), because the shallow cells resolved at $2^\circ$ cannot host two-thirds of the dissipation without distorting the open-ocean response. At this resolution the global-budget constraint is therefore the practical one, and the shelf target is a directional diagnostic whose quantitative use requires a grid that resolves the shelf seas.

\begin{table}[t]
\centering\footnotesize
\caption{Energy-penalty sweep with $\mu_D=\mu_{\mathrm{sh}}=\mu$ in Eq.~\eqref{eq:energypen} (hash CV, fold means). The $\mu=0$ row is the E1 fit.}
\label{tab:energy}
\setlength{\tabcolsep}{3.5pt}
\begin{tabular}{@{}cccccccc@{}}
\toprule
 & \multicolumn{3}{c}{\phys{}} & & \multicolumn{3}{c}{\nn{8}}\\
\cmidrule(lr){2-4}\cmidrule(lr){6-8}
$\mu$ & RMS & $D_{M_2}$ & shelf & & RMS & $D_{M_2}$ & shelf\\
 & [cm] & [TW] & [\%] & & [cm] & [TW] & [\%]\\
\midrule
0    & 37.7 & 2.55 & 21 & & \textbf{34.5} & 2.87 & 18\\
0.03 & 37.8 & 2.53 & 22 & & 35.0 & 2.78 & 22\\
0.1  & 37.8 & 2.49 & 25 & & 36.2 & 2.69 & 26\\
0.3  & 38.5 & 2.39 & 32 & & 40.3 & 2.52 & 38\\
1    & 41.3 & 2.25 & 48 & & 39.3 & \textbf{2.38} & \textbf{57}\\
\bottomrule
\end{tabular}
\end{table}

\begin{table}[t]
\centering\footnotesize
\caption{Energy-term ablation (hash CV). RMS is the fold mean of held-out error; open/coastal are pooled over 131/95 gauges. ``Budget'': $\mu_D=1,\mu_{\mathrm{sh}}=0$; ``budget + shelf'': $\mu_D=\mu_{\mathrm{sh}}=1$.}
\label{tab:ablation}
\setlength{\tabcolsep}{3pt}
\begin{adjustbox}{max width=\columnwidth}
\begin{tabular}{@{}llccccc@{}}
\toprule
Closure & Penalty & RMS [cm] & $D_{M_2}$ [TW] & Shelf [\%] & Open & Coastal\\
\midrule
\multirow{3}{*}{\phys{}} & none            & $37.7\pm5.5$ & 2.55 & $21\pm1$ & 21.7 & 54.6\\
                         & budget          & $37.8\pm5.5$ & 2.46 & $23\pm1$ & 21.5 & 54.8\\
                         & budget + shelf  & $41.3\pm5.5$ & 2.25 & $48\pm1$ & 26.3 & 58.2\\
\midrule
\multirow{3}{*}{\nn{8}}  & none            & $\mathbf{34.5}\pm3.3$ & 2.87 & $18\pm1$ & 20.6 & \textbf{49.3}\\
                         & budget          & $36.7\pm4.3$ & \textbf{2.42} & $19\pm3$ & \textbf{20.0} & 53.7\\
                         & budget + shelf  & $39.3\pm6.6$ & 2.38 & $\mathbf{57}\pm3$ & 23.3 & 56.4\\
\bottomrule
\end{tabular}
\end{adjustbox}
\end{table}

\subsection{Interpretability by Distillation (E6, Axis A4)}
\label{sec:res:distill}

We distil each fold's \nn{8} into closed-form drag laws by area-weighted least squares on its pre-activation $h_\theta(\bm\phi)$ and score the laws \emph{without refitting} (Table~\ref{tab:distill}). Low-order polynomials capture a growing share of the learned field, with mean $R^2$ of 0.28, 0.44 and 0.54 for degrees 1--3, and held-out error falls accordingly from 66.0 to 43.2 to 39.7\,cm. The tide solution is highly sensitive to the remaining part of the field, which is why the framework scores distilled laws through the full solve rather than by $R^2$ alone.

\begin{figure*}[t]
\centering
\figorbox{0.95\textwidth}{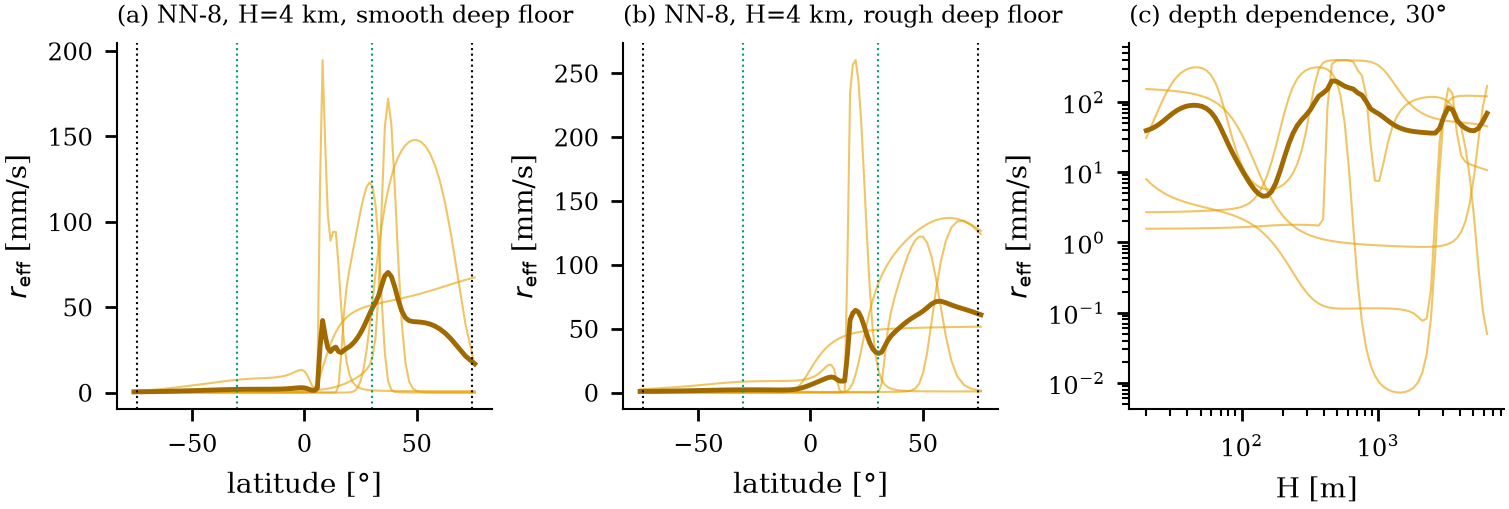}
\caption{Partial dependence of the learned drag $r_{\mathrm{eff}}$ of \nn{8} (thin lines: individual folds; thick line: mean). (a,b)~Latitude dependence at $H=4$\,km over a smooth (median slope) and a rough (90th-percentile slope) deep floor: near-zero drag in the south and large, fold-specific peaks in the north. (c)~Depth dependence at $30^\circ$, spanning four orders of magnitude across folds. Dotted lines: $M_2$ critical latitude (black) and $\pm30^\circ$ (green).}
\label{fig:pd}
\end{figure*}

The partial-dependence curves (Fig.~\ref{fig:pd}) show the structure that distillation has to capture. At 4\,km depth the learned drag is nearly zero ($1$--$3$\,mm\,s$^{-1}$) in every fold south of the equator. North of $\sim$5--20$^\circ$N it rises to 40--260\,mm\,s$^{-1}$, with each fold placing its peaks at different latitudes. The depth dependence at $30^\circ$ spans four orders of magnitude across folds. Folds with similar held-out skill thus reach different drag laws, which indicates that with $\sim$180 training gauges the data constrain the tidal response more tightly than the drag field itself. The framework provides the natural route to a usable closed form: because \poly{d} is itself a closure of the framework, a distilled law is refitted through the same adjoint pipeline with no new code.

\begin{table}[t]
\centering\footnotesize
\caption{Distillation of \nn{8} (114\,p) into polynomial laws $h(\bm\phi)$, no refit (hash CV).}
\label{tab:distill}
\setlength{\tabcolsep}{4pt}
\begin{tabular}{@{}lcccc@{}}
\toprule
Law & \#p & mean $R^2$ & held-out [cm] & worst fold [cm]\\
\midrule
degree 1 & 5  & 0.28 & 66.0 & 131.9\\
degree 2 & 11 & 0.44 & 43.2 & 51.1\\
degree 3 & 21 & 0.54 & 39.7 & 45.6\\
\midrule
\nn{8} (teacher)    & 114 & 1    & 34.5 & 40.3\\
\phys{} (reference) & 4   & --   & 37.7 & 44.6\\
\bottomrule
\end{tabular}
\end{table}

\subsection{Projection Stability under Sea-Level Rise (E7, Axis A5)}
\label{sec:res:slr}

\begin{figure*}[t]
\centering
\figorbox{0.95\textwidth}{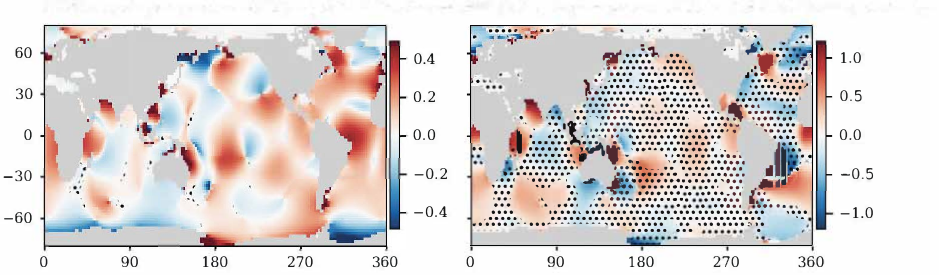}
\caption{Change in spring range per metre of uniform sea-level rise (fold-ensemble mean, cm\,m$^{-1}$); left: \phys{}, right: \nn{8}. Dots mark cells where the five fold models disagree in sign. The physical closure gives a coherent, reproducible pattern; for the neural closure the fold ensemble identifies the regions where projections depend on the training gauges.}
\label{fig:slrmap}
\end{figure*}

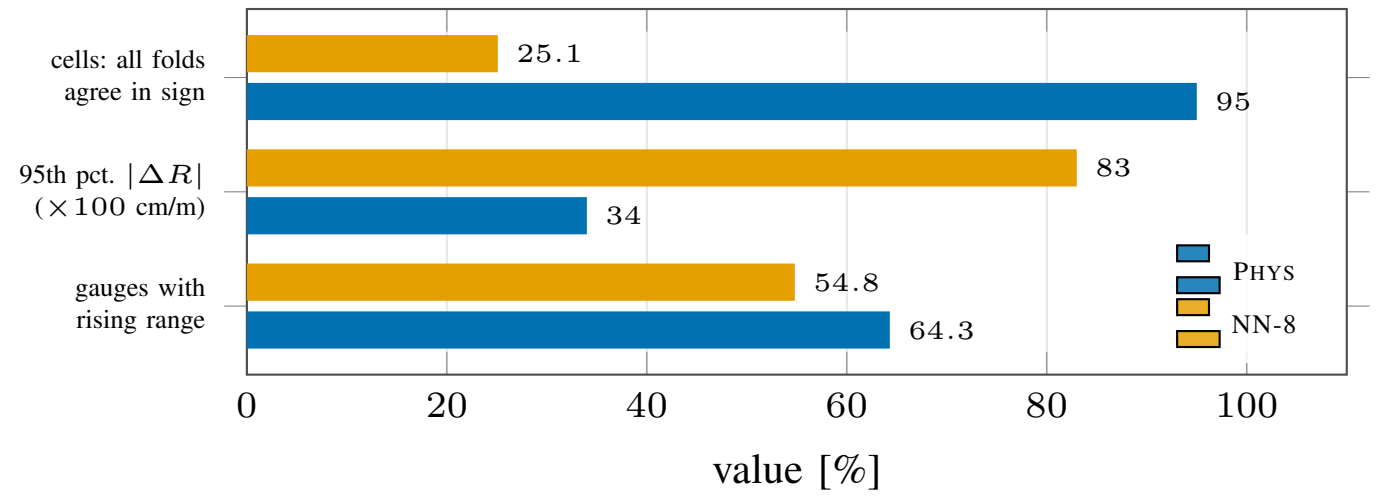
\begin{figure}[t]
\centering
\begin{tikzpicture}
\begin{axis}[hct, width=\columnwidth, height=4.0cm, xbar, bar width=7pt,
  xmin=0, xmax=110, ymin=0.4, ymax=3.6,
  ytick={1,2,3},
  yticklabels={{gauges with\\rising range},{95th pct.\ $|\Delta R|$\\($\times100$ cm/m)},{cells: all folds\\agree in sign}},
  yticklabel style={align=right, font=\tiny},
  xlabel={value [\%]}, xmajorgrids,
  legend style={at={(0.98,0.03)}, anchor=south east},
  nodes near coords, nodes near coords style={font=\tiny},
  every node near coord/.append style={anchor=west}]
\addplot[fill=cphys, draw=none] coordinates {(64.3,1)(34,2)(95.0,3)};
\addplot[fill=cnn, draw=none] coordinates {(54.8,1)(83,2)(25.1,3)};
\legend{\phys{}, \nn{8}}
\end{axis}
\end{tikzpicture}
\caption{Projection stability under $+1$\,m sea-level rise. The physical closure's response is reproducible across training folds; the fold ensemble quantifies how much of the neural closure's local response depends on the training gauges.}
\label{fig:slr}
\end{figure}

Projection is the end use of a forward tide model, and the framework evaluates it with the same solver. We deepen every wet cell by 1\,m, keep the coastline fixed, re-solve with each fold's $\theta^\star$ and compute the change in spring range $R=2(|M_2|+|S_2|)$ (Table~\ref{tab:slr}, Figs.~\ref{fig:slrmap} and~\ref{fig:slr}). The two closures agree on the global mean response ($+0.07$ and $+0.06$\,cm per m). The fold ensemble adds a measure of \emph{reproducibility}: for \phys{}, all five fold models agree on the sign of $\Delta R$ in 95\% of ocean cells, which makes its projected pattern robust to the choice of training gauges; for \nn{8}, they agree in 25\%, and its 95th-percentile local response is $2.4\times$ larger (0.83 vs.\ 0.34\,cm\,m$^{-1}$). This links directly to the fold-specific drag laws of Fig.~\ref{fig:pd}. Within the framework, fold-ensemble sign agreement costs only ten extra solves per closure and provides a direct, cheap measure of epistemic risk in projections.

\begin{table}[t]
\centering\footnotesize
\caption{Spring-range sensitivity to $+1$\,m uniform sea level (5-fold ensembles).}
\label{tab:slr}
\begin{tabular}{@{}lcc@{}}
\toprule
Metric & \phys{} & \nn{8}\\
\midrule
area-mean $\Delta R$ [cm/m]           & $+0.070$ & $+0.059$\\
gauge-mean $\Delta R$ [cm/m]          & $+0.158$ & $+0.109$\\
95th pct.\ $|\Delta R|$ [cm/m]        & 0.34 & 0.83\\
cells with all folds agreeing in sign & \textbf{95.0\%} & 25.1\%\\
gauges with increasing range          & 64.3\% & 54.8\%\\
\bottomrule
\end{tabular}
\end{table}

\subsection{Resolution Robustness (E9)}
\label{sec:res:res}

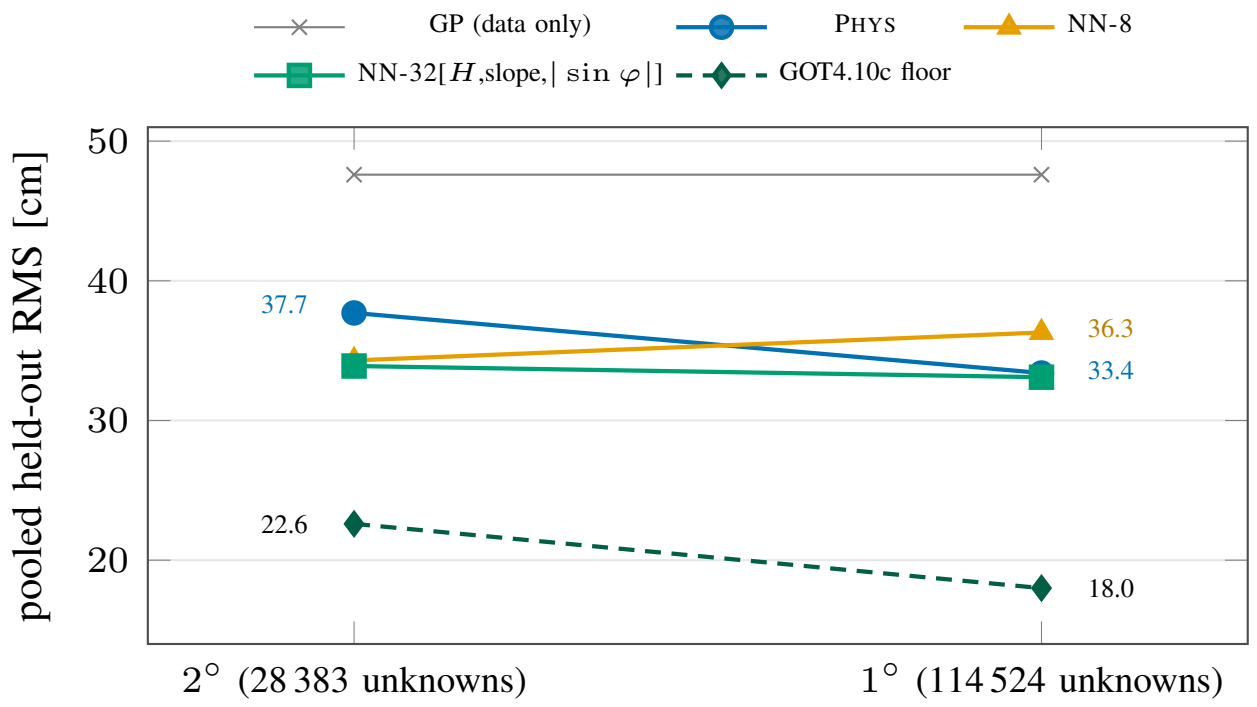
\begin{figure}[t]
\centering
\begin{tikzpicture}
\begin{axis}[hct, width=\columnwidth, height=5.0cm,
  xmin=-0.3, xmax=1.3, ymin=14, ymax=51,
  xtick={0,1}, xticklabels={{$2^\circ$ (28\,383 unknowns)},{$1^\circ$ (114\,524 unknowns)}},
  ylabel={pooled held-out RMS [cm]}, ymajorgrids,
  legend columns=3, legend style={at={(0.5,1.03)}, anchor=south}]
\addplot[gray, thin, mark=x] coordinates {(0,47.6)(1,47.6)};
\addlegendentry{GP (data only)}
\addplot[cphys, thick, mark=*] coordinates {(0,37.7)(1,33.4)};
\addlegendentry{\phys{}}
\addplot[cnn, thick, mark=triangle*] coordinates {(0,34.3)(1,36.3)};
\addlegendentry{\nn{8}}
\addplot[cgreen, thick, mark=square*] coordinates {(0,33.9)(1,33.1)};
\addlegendentry{\nn{32}[$H$,slope,$|\sin\varphi|$]}
\addplot[cgreen!60!black, thick, densely dashed, mark=diamond*, mark options={solid}] coordinates {(0,22.6)(1,18.0)};
\addlegendentry{GOT4.10c floor}
\node[font=\tiny, anchor=east] at (axis cs:-0.04,22.6) {22.6};
\node[font=\tiny, anchor=west] at (axis cs:1.04,18.0) {18.0};
\node[font=\tiny, anchor=east, cphys] at (axis cs:-0.04,38.3) {37.7};
\node[font=\tiny, anchor=west, cphys] at (axis cs:1.04,33.6) {33.4};
\node[font=\tiny, anchor=west, cnn!80!black] at (axis cs:1.04,36.6) {36.3};
\end{axis}
\end{tikzpicture}
\caption{Resolution test. Refining the grid lowers the representativeness floor by 4.6\,cm and the physical closure's error by 4.3\,cm. The signed-latitude network's $2^\circ$ lead is not retained (its $1^\circ$ difference to \phys{} is not significant); the hemispherically symmetric network is lowest on both grids and level with \phys{} at $1^\circ$.}
\label{fig:res}
\end{figure}

The diagnosis of Sec.~\ref{sec:res:maps} makes a testable prediction. If the signed-latitude correction stands in for coastal response that the $2^\circ$ grid does not resolve, its advantage should shrink when the grid is refined, whereas transferable skill should remain. Table~\ref{tab:res} and Fig.~\ref{fig:res} report the test.

\emph{(i) The guarantees carry over.} Over all fitted $1^\circ$ models, $|W-D|/W\le3.3\times10^{-16}$ and the solve residual is $\le2.0\times10^{-12}$. The cost per evaluation rises by 15\% from 4 to 1\,218 parameters (1.62 to 1.86\,s), and no component of the closure interface, the adjoint or the audit was changed.

\emph{(ii) Refinement removes representativeness error, not closure error.} The atlas floor falls from 22.6 to 18.0\,cm. The pooled error of \phys{} follows it from 37.7 to 33.4\,cm and improves in four of five folds, so its excess over the floor, taken as the difference of pooled scores, is about 15\,cm on both grids (15.1 and 15.4\,cm), in line with the paired estimate of E8 ($+15.2$\,cm). All three closures beat the GP in every fold at $1^\circ$, by 24--30\% in pooled error.

\emph{(iii) The signed-latitude lead is not retained.} The lead of \nn{8} over \phys{} at $2^\circ$ ($-3.4$\,cm [$-5.8,-1.0$]) becomes $+2.8$\,cm [$-2.5,+10.6$] at $1^\circ$, with 2/5 folds won. Its in-sample error keeps falling (32.7 to 27.5\,cm) while its generalization gap widens from $+1.9$ to $+7.8$\,cm, and a single fold contributes $+12.2$\,cm. The $1^\circ$ difference is not significant, so the supported statement is that the lead is no longer present, not that the ranking reverses.

\emph{(iv) The symmetric closure stays level.} \nn{32}[$H$, slope, $|\sin\varphi|$] has the lowest pooled error on both grids (33.9 and 33.1\,cm) and is statistically level with \phys{} at $1^\circ$ ($-0.3$\,cm [$-1.6,+1.2$], 3/5 folds). Its $2^\circ$ margin is not retained either: at $1^\circ$ both closures sit about 15\,cm above the floor, and no learned closure is significantly better than the 4-parameter law. This pattern is consistent with a large part of the random-CV gain at $2^\circ$ having compensated for unresolved coastal response, which the finer grid supplies directly. Because hyperparameters and the evaluation budget were left unchanged, the widening generalization gap of \nn{8} may also reflect overfitting on the finer grid; this test does not separate the two readings.

\emph{(v) Spatial-block validation pointed the same way.} For both learned closures the verdict of axis A1 at $2^\circ$ (Table~\ref{tab:block}: \nn{8} $+4.2$\,cm behind \phys{}, symmetric network $-0.3$\,cm) agrees in direction with the outcome of refinement. A single-resolution, random-CV study would have reported the $2^\circ$ gain without qualification; the audit flagged it before the finer grid was run.

\emph{(vi) Energy.} Unconstrained $M_2$ dissipation rises with resolution for every closure (\phys{} 2.55 to 2.92\,TW, \nn{8} 2.87 to 3.10\,TW, symmetric network 2.99 to 3.14\,TW), and the shelf share stays at 19--25\%. The shelf seas are thus still largely unresolved at $1^\circ$, and the global-budget constraint of Sec.~\ref{sec:res:energy} is the natural companion of finer-grid fits.

\begin{table*}[t]
\centering\footnotesize
\caption{Resolution test (hash CV, same 226 gauges and folds; RSS over $M_2,S_2,K_1,O_1$ in cm). The $2^\circ$ rows are refits with the E9 solver settings and reproduce Table~\ref{tab:cv} to within 0.1\,cm. $\Delta$ vs.\ \phys{}: pooled paired difference at the same resolution with 95\% bootstrap CI (negative = better). $|W{-}D|/W$: maximum over fitted models. Time per evaluation is measured with ten concurrent workers. GP pooled: 47.6\,cm.}
\label{tab:res}
\setlength{\tabcolsep}{4pt}
\begin{adjustbox}{max width=\textwidth}
\begin{tabular}{@{}llccccccccc@{}}
\toprule
Grid & Closure & In-sample & Held-out & Pooled & $\Delta$ vs.\ \phys{} [95\% CI] & Wins & $D_{M_2}$ [TW] & Shelf [\%] & $|W{-}D|/W$ & s/eval\\
\midrule
\multirow{3}{*}{$2^\circ$} & \phys{} & 37.2 & $37.7\pm5.5$ & 37.7 & -- & -- & 2.55 & 21 & $3.9\times10^{-16}$ & 0.25\\
 & \nn{8} & 32.7 & $34.5\pm3.4$ & 34.3 & $\mathbf{-3.4}\ [-5.8,-1.0]$ & 4/5 & 2.87 & 18 & $4.9\times10^{-16}$ & 0.26\\
 & \nn{32}[$H$, slope, $|\sin\varphi|$] & 31.7 & $34.0\pm4.4$ & \textbf{33.9} & $\mathbf{-3.8}\ [-6.3,-1.6]$ & 5/5 & 2.99 & 22 & $4.4\times10^{-16}$ & 0.27\\
\midrule
\multirow{3}{*}{$1^\circ$} & \phys{} & 31.6 & $32.9\pm8.9$ & 33.4 & -- & -- & 2.92 & 19 & $1.7\times10^{-16}$ & 1.62\\
 & \nn{8} & 27.5 & $35.3\pm10.9$ & 36.3 & $+2.8\ [-2.5,+10.6]$ & 2/5 & 3.10 & 25 & $3.3\times10^{-16}$ & 1.72\\
 & \nn{32}[$H$, slope, $|\sin\varphi|$] & 28.9 & $32.7\pm7.9$ & \textbf{33.1} & $-0.3\ [-1.6,+1.2]$ & 3/5 & 3.14 & 25 & $2.6\times10^{-16}$ & 1.86\\
\midrule
\multicolumn{2}{@{}l}{GOT4.10c at model cell (floor)} & \multicolumn{9}{l}{22.6 at $2^\circ$, \ 18.0 at $1^\circ$}\\
\bottomrule
\end{tabular}
\end{adjustbox}
\end{table*}

\subsection{Optimization Behaviour}
\label{sec:res:conv}
L-BFGS converges in 18--27 evaluations for \phys{}. The learnable closures use 64--76 evaluations of the fixed budget, which serves as implicit early stopping (Sec.~\ref{sec:objective}). A fit takes 12--18\,s for \phys{} and 32--58\,s for an MLP of any width. All 170 fits took 100\,min of optimizer time on one laptop CPU, which makes the full audit cheap enough to run routinely. At $1^\circ$ a fit takes about 30\,s for \phys{} (17--21 evaluations) and about 130\,s for an MLP (68--76 evaluations), with 1.6--1.9\,s per evaluation under ten concurrent workers.

\subsection{Admissibility Synthesis}
\label{sec:res:synth}

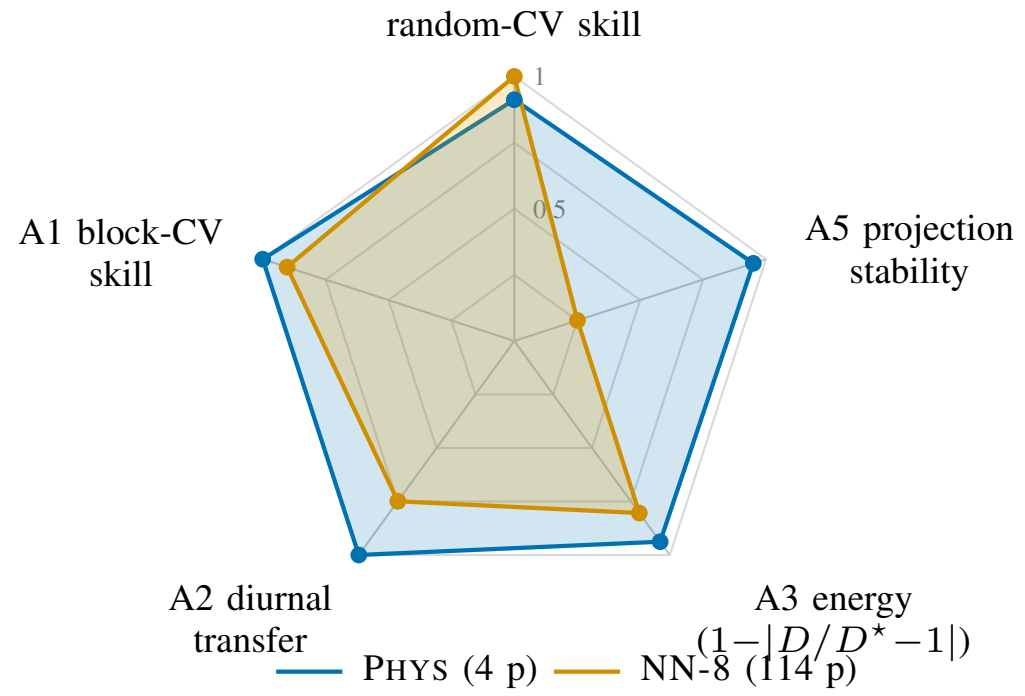
\begin{figure}[t]
\centering
\begin{tikzpicture}[scale=1.75, font=\scriptsize]
\foreach \r in {0.25,0.5,0.75,1.0}{
  \draw[gray!30] (90:\r) -- (162:\r) -- (234:\r) -- (306:\r) -- (18:\r) -- cycle;}
\foreach \a in {90,162,234,306,18}{ \draw[gray!45] (0,0) -- (\a:1); }
\node[align=center] at (90:1.2)  {random-CV skill};
\node[align=center, anchor=east] at (162:1.08) {A1 block-CV\\skill};
\node[align=center, anchor=north east] at (234:1.05) {A2 diurnal\\transfer};
\node[align=center, anchor=north west] at (306:1.05) {A3 energy\\($1{-}|D/D^\star{-}1|$)};
\node[align=center, anchor=west] at (18:1.08) {A5 projection\\stability};
\node[gray, font=\tiny, right] at (90:0.5) {0.5};
\node[gray, font=\tiny, right] at (90:1.0) {1};
\draw[cphys, thick, fill=cphys, fill opacity=0.18]
  (90:0.912) -- (162:1.0) -- (234:1.0) -- (306:0.938) -- (18:0.950) -- cycle;
\draw[cnn!90!black, thick, fill=cnn, fill opacity=0.22]
  (90:1.0) -- (162:0.903) -- (234:0.749) -- (306:0.804) -- (18:0.251) -- cycle;
\foreach \a/\r in {90/0.912,162/1.0,234/1.0,306/0.938,18/0.950}{\fill[cphys] (\a:\r) circle (0.9pt);}
\foreach \a/\r in {90/1.0,162/0.903,234/0.749,306/0.804,18/0.251}{\fill[cnn!90!black] (\a:\r) circle (0.9pt);}
\draw[cphys, thick] (-0.9,-1.25) -- ++(0.25,0) node[right, black]{\phys{} (4 p)};
\draw[cnn!90!black, thick] (0.15,-1.25) -- ++(0.25,0) node[right, black]{\nn{8} (114 p)};
\end{tikzpicture}
\caption{Admissibility profile (higher is better). Skill axes are normalized as best/value (pooled RMS; diurnal axis: RSS of $K_1$ and $O_1$ under $M_2$-only fitting). The energy axis is $1-|D_{M_2}/2.4\,\mathrm{TW}-1|$; projection stability is the fraction of cells where the fold ensemble agrees in sign. Axis A4 (interpretability) is qualitative and is reported in Table~\ref{tab:scorecard}. All values are at $2^\circ$.}
\label{fig:radar}
\end{figure}

Fig.~\ref{fig:radar} and Table~\ref{tab:scorecard} combine the axes for the two reference closures. At $2^\circ$ the neural closure leads on random-CV skill, is about 1\,cm lower at open-ocean gauges under both validation designs, agrees more closely with the atlas in deep water, and responds well to energy constraints, where the global budget alone brings it onto the 2.4\,TW target at a fold-mean error about 1\,cm below \phys{}. Its random-CV lead is not retained at $1^\circ$ (Sec.~\ref{sec:res:res}), in line with what spatial-block validation had indicated. The physical closure leads on spatial extrapolation at coastal gauges, frequency transfer, unconstrained energy admissibility, interpretability and projection stability, at the same computational cost. The scorecard is the framework's end product: a complete, quantitative profile of every closure, of which a skill-only evaluation would report a single row. The same profile defines the promotion path for the hemispherically symmetric network: it has cleared axis A1 (Table~\ref{tab:block}) and the resolution test (Table~\ref{tab:res}), whereas its width-8 counterpart transfers poorly to the diurnal band under $M_2$-only fitting (Table~\ref{tab:transfer}), so A2, A4 and A5 under joint fitting are the remaining audits before it can replace \phys{}.

\begin{table}[t]
\centering\footnotesize
\caption{Admissibility scorecard (green: better closure on that axis). Rows are at $2^\circ$ unless stated. Strata and atlas rows use the 221 benchmark gauges.}
\label{tab:scorecard}
\setlength{\tabcolsep}{2.5pt}
\begin{adjustbox}{max width=\columnwidth}
\begin{tabular}{@{}llcc@{}}
\toprule
Axis & Metric & \phys{} & \nn{8}\\
\midrule
Skill (random CV) & pooled RMS [cm] & 37.7 & \cellcolor{good}34.4\\
A1 Spatial extrap. & pooled block RMS [cm] & \cellcolor{good}39.0 & 43.2\\
A1 open gauges & block RMS [cm] & 22.4 & \cellcolor{good}21.6\\
A1 coastal gauges & block RMS [cm] & \cellcolor{good}57.2 & 65.2\\
Atlas (deep field) & RSS vs.\ GOT4.10c [cm] & 15.4 & \cellcolor{good}14.0\\
Resolution ($1^\circ$, random CV) & pooled RMS [cm] & \cellcolor{good}33.4 & 36.3 (n.s.)\\
A2 Freq.\ transfer & $K_1$ / $O_1$ [cm] & \cellcolor{good}8.4 / 6.8 & 10.3 / 10.1\\
A3 Energy ($\mu{=}0$) & $D_{M_2}$ [TW] & \cellcolor{good}2.55 & 2.87\\
A3 Energy (budget) & RMS [cm] / $D_{M_2}$ [TW] & 37.8 / 2.46 & \cellcolor{good}36.7 / 2.42\\
A3 Energy ($\mu{=}1$) & RMS / shelf share & 41.3 / 48\% & \cellcolor{good}39.3 / 57\%\\
A4 Interpretability & closed form? & \cellcolor{good}4 named coeffs. & $R^2\le0.54$\\
A5 Projection & fold sign agreement & \cellcolor{good}95\% & 25\%\\
Cost & s per evaluation & 0.65 & 0.66\\
\bottomrule
\end{tabular}
\end{adjustbox}
\end{table}

\section{Conclusion}
\label{sec:conclusion}
We presented \hct{}, a hard-constrained differentiable framework in which a learnable closure predicts only the drag field and an SAL factor, while the discretized frequency-domain tidal equations are solved to round-off. A strictly positive damping parameterization, an energy-consistent C-grid and an implicit adjoint make every parameter vector admissible, close the discrete energy budget to round-off and keep the gradient cost almost independent of closure size. The same solve powers a five-axis admissibility audit, an atlas benchmark and a grid-refinement test that distinguish closures which random cross-validation ranks as equal or better.

On 226 UHSLC gauges, every constrained closure beats data-only interpolation by 17--29\% in total error and by 33--55\% in RMS excess over the $2^\circ$ representativeness floor, which itself equals 60\% of the physical closure's RMS error. The physical closure's margin grows under spatial extrapolation, most strongly in the open ocean. The audit splits a 114-parameter neural closure's random-CV lead into two parts: an open-ocean improvement that transfers across regions and is supported by closer agreement with the altimetric atlas in deep water, and a hemispherically asymmetric drag correction concentrated in the latitude bands of the northern shelf seas, which improves coastal gauges only within the region it was trained on. The framework then converts this diagnosis into more robust closures. A hemispherically symmetric network is the best closure under random CV and statistically level with the physical closure under spatial blocking. Refinement to $1^\circ$ is consistent with the audit. The representativeness floor falls by 4.6\,cm and the physical closure's error by 4.3\,cm to 33.4\,cm, the signed-latitude network's lead is no longer present, the symmetric network stays level with the physical closure, and the energy identity still holds to round-off at 114\,524 unknowns. At $2^\circ$, a global energy constraint brings the neural closure onto the altimetric budget target while its fold-mean held-out error stays about 1\,cm below that of the physical law, whereas the shelf-share target is usable at this resolution only as a directional diagnostic. The resulting decision procedure is simple: use the compact physical closure as the baseline, use the neural closure as a diagnostic of deficient physics, and promote learned closures once their gains hold on all five audit axes.

Because the solver, the constraint layer, the adjoint and the audit communicate only through the closure interface, \hct{} moved from $2^\circ$ to $1^\circ$ without changes to any of them, and it carries over in the same way to finer grids with Krylov solvers, to quadratic and wave-drag closures, to multi-year gauge and altimetry records, and to additional reference atlases such as the $1/8^\circ$ EOT20~\cite{hartdavis2021}. These extensions will lower the representativeness floor further, resolve the shelf seas that neither grid yet resolves, and let the framework test the closures it discovers against the full global tidal energy budget.

\section*{Code and Data Availability}

The complete HCTide implementation is openly available at
\url{https://github.com/BorisKriuk/HCTide} under the MIT
License. ETOPO1, UHSLC and GOT4.10c data are obtained from their providers and are subject to their terms.


\end{document}